\documentclass[runningheads]{llncs}

\usepackage[mobile]{eccv}

\usepackage{eccvabbrv}

\usepackage{colortbl}
\usepackage{hhline}
\usepackage{multirow}
\usepackage{graphicx}
\usepackage{booktabs}
\usepackage{microtype}
\usepackage{subcaption}

\usepackage[accsupp]{axessibility}  

\usepackage[pagebackref,breaklinks,colorlinks,citecolor=eccvblue]{hyperref}

\usepackage{amsmath}
\usepackage{mathtools}
\usepackage{overpic}
\usepackage{color}
\usepackage{listings} 
\usepackage[skip=.2em]{caption} 
\usepackage{xcolor}
\usepackage{currfile}
\usepackage{cancel}
\usepackage{float}
\usepackage{rotating}
\usepackage{booktabs}
\usepackage{comment}
\usepackage{xspace}
\usepackage{bm}

\usepackage{tabularx}
\usepackage{etoolbox,siunitx}
\robustify\bfseries
\usepackage{cuted} 

\usepackage{currfile} 

\usepackage{soul}
\setuldepth{foobar}

\usepackage{caption} 

\usepackage[rightcaption]{sidecap}  
\sidecaptionvpos{figure}{t} 
\definecolor{turquoise}{cmyk}{0.65,0,0.1,0.3}
\definecolor{purple}{rgb}{0.65,0,0.65}
\definecolor{dark_green}{rgb}{0, 0.5, 0}
\definecolor{orange}{rgb}{0.9, 0.6, 0.2}
\definecolor{red}{rgb}{0.8, 0.2, 0.2}
\definecolor{darkgray}{rgb}{0.5, 0.5, 0.5}
\definecolor{darkred}{rgb}{0.6, 0.1, 0.05}
\definecolor{blueish}{rgb}{0.0, 0.3, .6}
\definecolor{light_gray}{rgb}{0.7, 0.7, .7}
\definecolor{pink}{rgb}{1, 0, 1}
\definecolor{greyblue}{rgb}{0.25, 0.25, 1}
\definecolor{BrightGreen}{rgb}{0.196, 0.7, 0.196}

\definecolor{tabgreen}{HTML}{81C784}
\definecolor{tabyellow}{HTML}{AED581}
\definecolor{tabyellowlight}{HTML}{DCE775}
\definecolor{Gray}{gray}{0.95}

\usepackage[shortlabels,inline]{enumitem}
\setlist[itemize]{noitemsep,leftmargin=*,topsep=0em}
\setlist[enumerate]{noitemsep,leftmargin=*,topsep=0em}

\usepackage{bbm}
\usepackage{lipsum}
\usepackage{blindtext}

\newcommand{\SupplementaryMaterial}[1]{{supplementary material}}

\usepackage{enumitem}
\setlist[itemize]{noitemsep,leftmargin=*,topsep=0em}
\setlist[enumerate]{noitemsep,leftmargin=*,topsep=0em}

\newcommand{\MethodName}{\texttt{360Anything}\xspace}
\newcommand{\Website}{https://360anything.github.io}
\newcommand{\PaperTitle}{Geometry-Free Lifting of \\ Images and Videos to $360^\circ$}

\newcommand{\inpers}{\mathrm{X_{pers}}}
\newcommand{\inpano}{\mathrm{X_{equi}^{proj}}}
\newcommand{\outpano}{\mathrm{Y_{equi}}}
\newcommand{\outpanonoisy}{\mathrm{Y_{equi}^t}}
\newcommand{\inperstoken}{\mathrm{x_{pers}}}

\newcommand{\outpanotoken}{\mathrm{y_{equi}}}
\newcommand{\outpanotokennoisy}{\mathrm{y_{equi}^t}}
\newcommand{\textembs}{\bm{e}}

\newcommand{\param}{\bm{\theta}}
\newcommand{\latentencoder}{\mathcal{E}}
\newcommand{\latentdecoder}{\mathcal{D}}
\newcommand{\denoiser}{\mathcal{G}_{\param}}

\newcommand{\StdNormal}{\mathcal{N}(\bm{0}, \bm{I})}

\newcommand{\heading}[1]{\vspace{.25em}\noindent\textbf{#1}}

\usepackage[capitalize]{cleveref}
\crefname{section}{Sec.}{Secs.}
\Crefname{section}{Section}{Sections}
\crefname{appendix}{App.}{Apps.}
\Crefname{appendix}{Appendix}{Appendices}
\Crefname{table}{Table}{Tables}
\crefname{table}{Tab.}{Tabs.}
\Crefname{figure}{Figure}{Figures}
\crefname{figure}{Fig.}{Figs.}
\Crefname{algorithm}{Algorithm}{Aalgorithms}
\crefname{algorithm}{Algo.}{Algos.}

\newlength\pagetopmargin
\newlength\figcapmargin
\newlength\figmargin
\newlength\tablecapmargin
\newlength\tablemargin

\renewcommand{\arraystretch}{1.1}  

\usepackage{orcidlink}

\begin{document}

\title{\MethodName: \PaperTitle}

\titlerunning{\MethodName: Geometry-Free Lifting of Images and Videos to $360^\circ$}

\author{
Ziyi Wu\inst{1,3,\dagger} \and 
Daniel Watson\inst{1} \and 
Andrea Tagliasacchi\inst{2,3,\dagger} \and \\ 
David J. Fleet\inst{1,3,\dagger} \and 
Marcus A. Brubaker\inst{1} \and 
Saurabh Saxena\inst{1} 
}

\authorrunning{Z.~Wu et al.}

\institute{
$^{1}$ Google DeepMind
$^{2}$ Simon Fraser University
$^{3}$ University of Toronto
}

\maketitle

{
  \renewcommand{\thefootnote}{\fnsymbol{footnote}}
  \footnotetext[4]{Work done at Google DeepMind.}
}

\begin{abstract}

Lifting perspective images and videos to 360$^\circ\!$ panoramas enables immersive 3D world generation.
Existing approaches often rely on explicit geometric alignment between the perspective and the equirectangular projection (ERP) space.
Yet, this requires known camera metadata, obscuring the application to in-the-wild data where such calibration is typically absent or noisy.
We propose \MethodName, a geometry-free framework built upon pre-trained diffusion transformers.
By treating the perspective input and the panorama target simply as token sequences, \MethodName learns the perspective-to-equirectangular mapping in a purely data-driven way, eliminating the need for camera information.
Our approach achieves state-of-the-art performance on both image and video perspective-to-360$^\circ\!$ generation, outperforming prior works that use ground-truth camera information.
We also trace the root cause of the seam artifacts at ERP boundaries to zero-padding in the VAE encoder, and introduce \emph{Circular Latent Encoding} to facilitate seamless generation.
Finally, we show competitive results in zero-shot camera FoV and orientation estimation benchmarks, demonstrating \MethodName's deep geometric understanding and broader utility in computer vision tasks.
Additional results are available at \url{\Website}.


\end{abstract}

\begin{figure}[t]
  \vspace{-0.5mm}
  \centering
  \begin{subfigure}[b]{\textwidth}
    \centering
    \includegraphics[width=\textwidth]{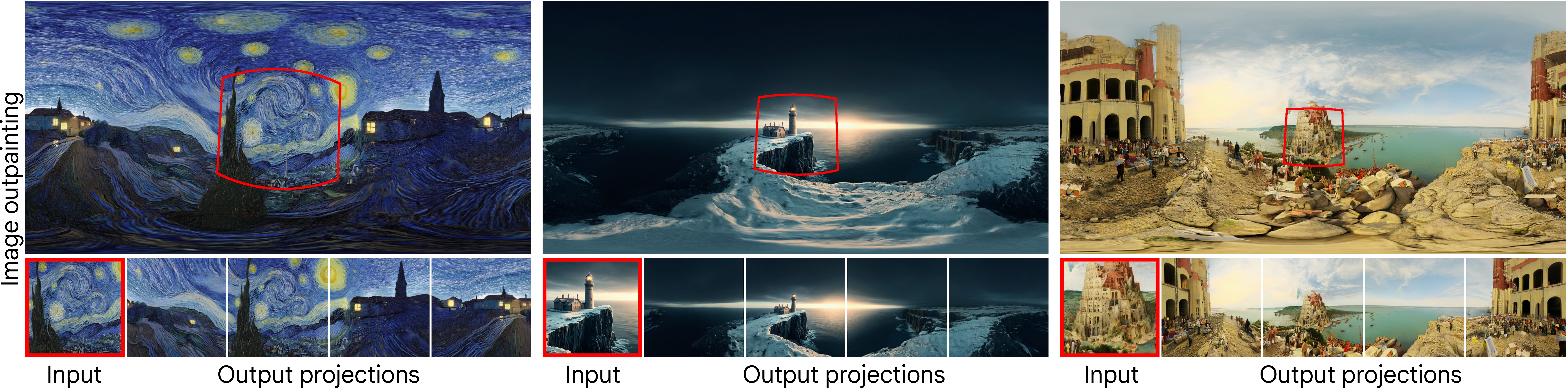}
  \end{subfigure}
  \\
  \vspace{0mm}
  \begin{subfigure}[b]{\textwidth}
    \centering
    \includegraphics[width=\textwidth]{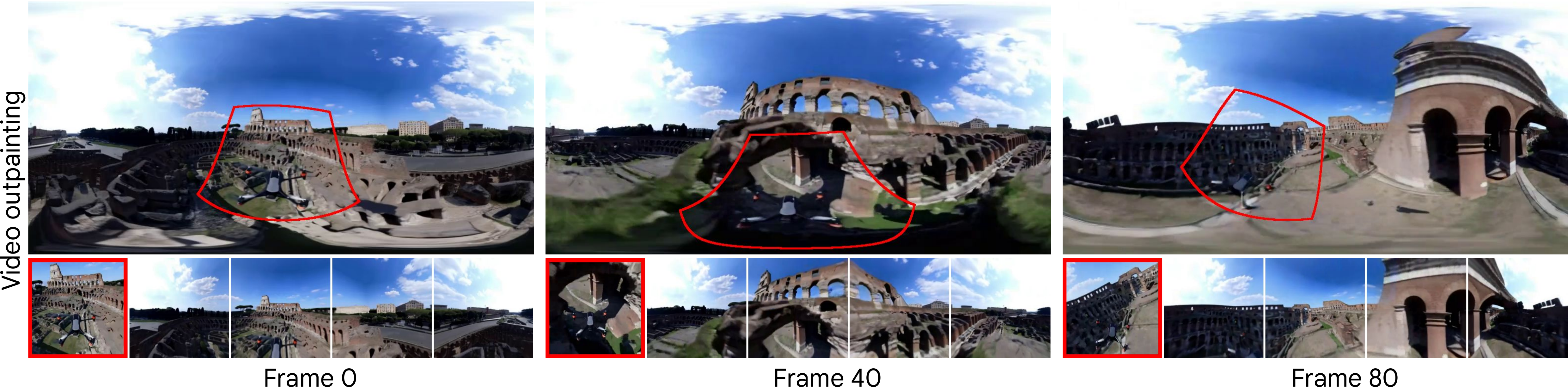}
  \end{subfigure}
  \\
  \vspace{0mm}
  \begin{subfigure}[b]{\textwidth}
    \centering
    \includegraphics[width=\textwidth]{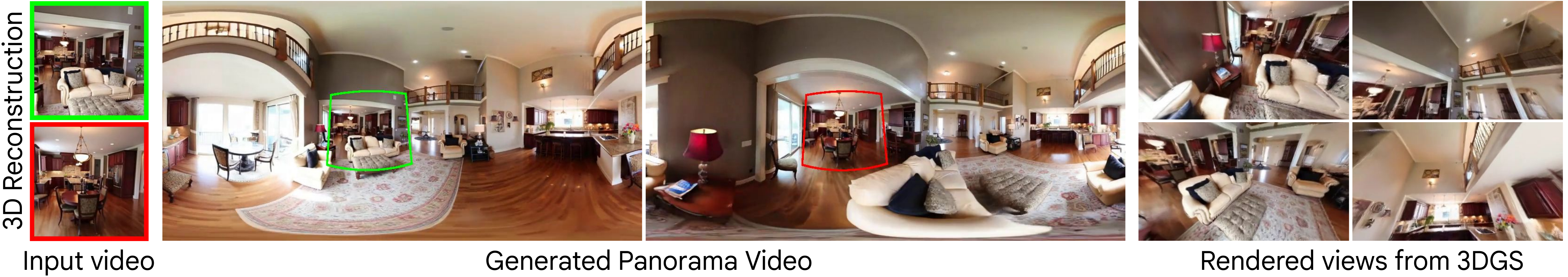}
  \end{subfigure}
  \vspace{\figcapmargin}
  \caption{
  \MethodName lifts arbitrary perspective images (row 1) and videos (row 2) to seamless, gravity-aligned 360$^\circ\!$ panoramas.
  Model inputs and their projected regions are highlighted in {\color{red} \textbf{red}} or {\color{green} \textbf{green}}.
  Below each panorama, we show four perspective projections facing left, front, right, and back.
  Without using explicit camera information, \MethodName handles images with varying Field-of-View and videos with large object and camera motion.
  The generated consistent panoramas enable 3D scene reconstruction via 3D Gaussian Splatting (row 3).
  Please see our \href{\Website}{project page} for results in 360$^\circ\!$ viewers.
  }
  \label{fig:teaser}
  \vspace{\figmargin}
  \vspace{-1.5mm}
\end{figure}

\section{Introduction}
\label{sec:intro}

Generating photorealistic 3D worlds is an exciting and challenging frontier in generative modeling, offering transformative potential across robotics, AR/VR, and gaming.
Recent years have witnessed significant advancements in this domain~\cite{Schwarz2025worlds, HunyuanWorld, MetaWorldGen, Marble}, largely propelled by the dramatic progress in visual generative models~\cite{Wan2.1, FLUX, LDM, Sora, GoogleVeo}.
However, standard generators produce \textit{perspective} views, capturing a narrow view of the physical world, and limiting their utility in creating fully immersive 3D worlds.
This limitation has spurred significant interest in 360$^\circ\!$ generative models~\cite{LDM3D, StitchDiffusion, 360DVDVideo, PanoDiTVideo}, especially those for lifting perspective imagery to omnidirectional 360$^\circ\!$ panoramas~\cite{Diffusion360, CubeDiff, Argus, Imagine360}.

Despite recent progress, current approaches for perspective-to-panorama generation lack the robustness needed for ``in-the-wild'' inputs.
To bridge the gap between perspective and panoramic spaces, prior works often rely on strong geometric inductive biases, such as \textit{explicitly} projecting the perspective input to the target Equirectangular Projection (ERP) space to provide an aligned conditioning signal~\cite{PanoDiffusion, Argus, Imagine360, PanoDecouple}.
However, this strategy requires known camera metadata, such as Field-of-View (FoV) and camera pose (yaw, pitch, roll) \cite{Argus}.
As a consequence, these models struggle with inputs where such metadata is absent, or are brittle when relying on noisy external estimators.

We posit that explicit geometric alignment is unnecessary for panorama generation.
Instead, with sufficient data, a general-purpose architecture should be able to learn these relationships from data.
To this end, our proposed framework utilizes a diffusion transformer (DiT)~\cite{DiT}, and treats the perspective input and target panorama simply as token sequences.
With attention on the concatenated sequence, the model learns their geometric relationship.
This enables the model to effectively ``place'' the perspective input onto the 360$^\circ\!$ canvas and synthesize the remaining context, handling varying FoVs and camera poses as shown in \Cref{fig:teaser}.
Our pipeline thus eliminates the camera estimation step and makes the task fully end-to-end, which enjoys the benefit of scaling up model and data.

Beyond an end-to-end framework, we also address the common issue of seam artifacts at the boundary of a generated ERP image.
Existing works rely on inference time tricks such as rotation augmentations to mitigate visible seams~\cite{360DVDVideo, Argus, PanoDiffusion}.
In contrast, we identify and eliminate the root cause of these artifacts during the training stage itself.
Modern diffusion models often operate in the latent space of a convolution-based VAE~\cite{LDM}, and VAEs utilize zero-padding in convolutional layers~\cite{CNNPosInfo}.
This introduces boundary artifacts in the latent representation of panorama data, which leads to seams in the generated panoramas.
We propose a simple solution that uses \emph{circular padding} when encoding VAE latents. This ensures that the latent representation has circular continuity, thereby eliminating the root cause of seam artifacts.

In summary, this work makes the following contributions:
\begin{enumerate}
\itemsep 0.1cm
    \item We propose \MethodName, a novel DiT-based architecture for ``in-the-wild'' perspective to canonical panorama generation that implicitly infers camera intrinsics and extrinsics, eliminating the need for camera calibration.
    \item We identify VAE latent encoding as the root cause of seam artifacts in panorama generation and propose a simple remedy that mitigates the issue.
    \item Despite not using camera metadata, \MethodName achieves state-of-the-art performance for panoramic image and video generation, outperforming baselines that have access to extra camera information.
    \item We evaluate the accuracy of our estimated FoV and camera poses, showing competitive results against supervised baselines.
    Furthermore, we can reconstruct consistent 3D scenes from our generated panoramic videos.
\end{enumerate}

\section{Related Work}
\label{sec:related}

\heading{Panorama Image Generation.}
Early approaches used GANs~\cite{GANs, Pers2Pano1, Pers2Pano2, Pers2Pano3, AOG-Net, CubeGAN, EpipolarGAN} or autoregressive generators~\cite{VQGAN, Text2Light, OmniDreamer, PanoLlama}.
Recent methods have switched to diffusion models~\cite{DDPM, LDM} due to their state-of-the-art performance.
One line of work generates panoramas from text prompts~\cite{LDM3D, StitchDiffusion, Diffusion360, WhatMakesPanoGen}.
They often design better panorama representations~\cite{TanDiT, SphereDiff} or panorama-aware operations~\cite{DiffPano, PanFusion} to reuse knowledge in pre-trained perspective generators.
Closer to \MethodName are methods that outpaint panoramas from narrow field-of-view images~\cite{Pers2Pano5, OmniPanoGen2, MaskedPanoGen, PanoFree}.
The majority of them project the conditioning perspective image to ERP space to be pixel-aligned with the target panorama, and then execute diffusion in ERP space~\cite{Pers2Pano4, PanoDiffusion}.
To handle the geometric properties of panoramas, they often inject strong inductive bias such as spherical convolutions~\cite{SphericalPer2Pano1, SphericalPer2Pano2} and dual branch architectures~\cite{PanoDecouple}.
A few works explore the cubemap representation to eliminate large distortions inherent in the ERP~\cite{CubeDiff, DreamCube}.
However, existing methods either require known camera information to perform the projection, or assume the conditioning image has a fixed viewpoint and FoV.
In contrast, our method treats the problem as a sequence-to-sequence task learned directly from data.

\heading{Panorama Video Generation.}
Some works directly fine-tune pre-trained video generators to produce panorama videos from text~\cite{360DVDVideo, PanoDiTVideo, DynamicScaler, PanoWan}.
This paper instead tackles the task of panoramic outpainting from perspective videos~\cite{VidPanos, VideoPanda, LuxDiT}.
Imagine360~\cite{Imagine360} duplicates the denoising U-Net in AnimateDiff~\cite{AnimateDiff} to process panorama and perspective views separately, connected by spherical attention for information exchange.
ViewPoint~\cite{ViewPointPanoVideo} proposes an improved cubemap representation to reduce the geometric distortion of ERP.
Argus~\cite{Argus} further scales up training data to unconstrained YouTube videos~\cite{360-1M}.
Nevertheless, these methods 
\begin{enumerate*}[label=(\roman*)] 
\item rely on \textit{external tools} to estimate camera metadata of the conditioning video, and 
\item use \textit{inference tricks} to eliminate seams in the generation. 
\end{enumerate*}
In contrast, \MethodName learns to pose the input video in 4D space, and remove seams by identifying its root cause and adjusting the architecture.

\heading{Prior-Free Learning with Transformers.}
Recently, Transformers have dominated tasks that previously relied on inductive bias, including image generation~\cite{OmniGen, DiT}, editing~\cite{Qwen-Image, FLUX-Kontext}, and 3D understanding~\cite{DUSt3R, GeoFreeNVS}.
While the majority of panorama generation methods still leverage a U-Net based architecture~\cite{CubeDiff, Argus}, we identify this as a major limiting factor for the field.
\MethodName instead runs Transformers on a sequence of tokens without any geometric prior, while achieving state-of-the-art results across multiple tasks.

\section{Method}
\label{sec:method}

\heading{Task formulation.}
Given a perspective video with $T$ frames, $\inpers \in \mathbb{R}^{T \times h \times w \times 3}$ (we treat image as a special case with $T{=}1$) and a caption $\textembs$, our goal is to outpaint a 360$^\circ\!$ panoramic video $\outpano \in \mathbb{R}^{T \times H \times W \times 3}$.
In this work, we represent panorama data in the Equirectangular Projection (ERP) space.

\begin{figure*}[!t]
\vspace{\pagetopmargin}
    \centering
    \includegraphics[width=\linewidth]{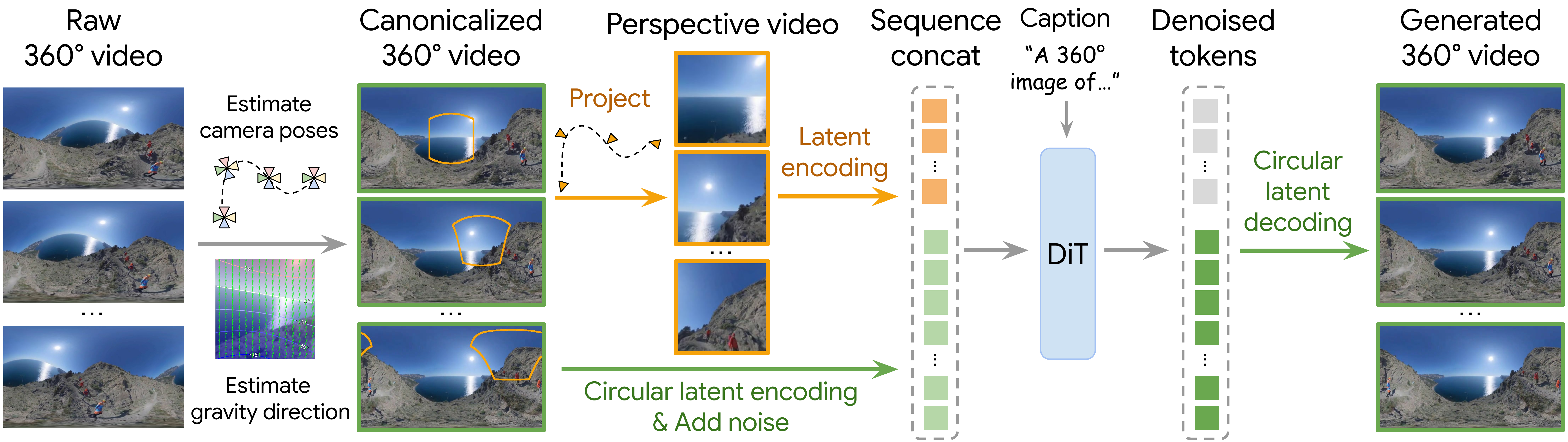}
    \vspace{\figcapmargin}
    \caption{
    \textbf{\MethodName pipeline.}
    Given a raw 360$^\circ\!$ training video with arbitrary camera orientations, we first estimate per-frame camera poses and rotate frames to align with the first frame.
    We then estimate the video's gravity direction and align it with the vertical axis.
    With such a \emph{canonicalized} 360$^\circ\!$ video, we project it to a perspective video using randomly sampled camera intrinsics and poses (\cref{subsec:pano-video-gen}). 
    We then encode both the conditioning and target videos to latent tokens.
    Critically, we employ \emph{Circular Latent Encoding} for the target 360$^\circ\!$ video to avoid seam artifacts in the latent representation.
    The conditioning tokens ({\color{orange} \textbf{orange}}) and noisy target tokens ({\color{BrightGreen} \textbf{green}}) are concatenated along the \emph{sequence dimension} and fed into a diffusion transformer~(DiT).
    The denoised tokens can be decoded back to a 360$^\circ\!$ video via circular latent decoding.
    }
\vspace{\figmargin}
\label{fig:method}
\end{figure*}

\heading{Overview.}
Our method builds upon pre-trained latent diffusion transformers~\cite{DiT} (\cref{subsec:background}).
We leverage a simple sequence concatenation approach to learn the perspective-to-equirectangular mapping and generate panoramas in a gravity-aligned canonical space (\cref{subsec:geofree-panogen}).
Finally, we address the seam artifacts of generated panoramas by analyzing the panoramic latent space (\cref{subsec:seamless-pano}).
The overall architecture of \MethodName is shown in \Cref{fig:method}.

\subsection{Background}
\label{subsec:background}

We adopt the flow matching framework~\cite{FlowMatching, RFSampler}, which learns a denoiser $\denoiser$ that maps from the standard normal distribution $\bm{\epsilon} \sim \StdNormal$ to the distribution of panorama data $\outpano \sim p_{\text{data}}$.
The forward diffusion process adds noise to clean data to obtain a noisy input $\outpanonoisy$ at time $t \in [0,1]$; i.e., $\outpanonoisy = (1-t) \outpano + t \bm{\epsilon}$.
The denoiser $\denoiser$, implemented as a neural network parameterized by $\param$, is trained to reverse this process with the following objective:
\begin{equation}
\min_{\param}\ {\mathbb{E}}_{ t \sim p(t), \outpano \sim p_{\text{data}}, \bm{\epsilon} \sim \StdNormal } \| (\bm{\epsilon} - \outpano) - \denoiser(\outpanonoisy, t, \bm{c}) \|^2,
\end{equation}
where $p(t)$ is the distribution of noise levels~\cite{SD3} and $\bm{c}$ refers to the auxiliary conditioning inputs, which in our case are the caption and the perspective input.
We implement the denoiser $\denoiser$ as a diffusion transformer (DiT)~\cite{DiT, FLUX, Wan2.1}.

To generate samples at a high resolution, modern diffusion models typically operate in the latent space of a pre-trained convolution-based VAE~\cite{LDM}.
Following this practice, we encode panorama data $\outpano$ to a latent representation $\outpanotoken$ via an encoder $\latentencoder$, which can be decoded back to the pixel space with a decoder $\latentdecoder$:
\begin{equation}
    \outpanotoken = \latentencoder(\outpano),\ \ \mathrm{\hat{Y}_{equi}} = \latentdecoder(\outpanotoken).
\end{equation}
The latent representation $\outpanotoken$ is then patchified and flattened into a 1D sequence of tokens that is provided as input to the DiT.

\subsection{Geometry-Free Scalable Panorama Generation}
\label{subsec:geofree-panogen}

A core challenge in perspective-to-panorama generation lies in finding an effective solution for conditioning the model on the perspective input $\inpers$.
Prior works~\cite{Argus, Imagine360, Pers2Pano4, PanoDiffusion} typically project $\inpers$ into the ERP space to obtain $\inpano$, which is \textit{pixel-aligned} with the generation target $\outpano$.
Then, they concatenate the latent of $\inpano$ and the noisy latent $\outpanotokennoisy$ \emph{channel-wise} as the input to the diffusion model.
This approach imposes a strong geometric inductive bias by explicitly localizing the perspective input $\inpers$ on the panorama output $\outpano$, drastically simplifying the task to image outpainting.
However, pixel-aligned perspective-to-panorama projection requires precise camera Field-of-View (FoV) and orientation estimates~\cite{Argus}.
For in-the-wild test data, this information is generally unavailable, and existing estimation methods can be noisy~\cite{MoGe, GeoCalib, MegaSaM}, resulting in accumulated errors and suboptimal performance.
Consequently, when off-the-shelf camera estimators fail, channel-concatenation approaches break down completely due to the reliance on pixel-aligned conditioning; see Appendix~\ref{app-subset:channel-concat-failure}.

\heading{Sequence concatenation.}
We propose to relax this constraint by treating geometric alignment as a task we can learn from data.
Instead of enforcing spatial correspondence via projection into the ERP space, we employ a simple \emph{sequence concatenation} mechanism inspired by recent image editing models~\cite{Qwen-Image, FLUX-Kontext}.
We directly encode the perspective input to latents: $\inperstoken = \latentencoder(\inpers)$, and append it to the noisy latents $\outpanotokennoisy$ as the DiT input: $\mathrm{Concat([\inperstoken, \outpanotokennoisy])}$.
The DiT thus runs global self-attention on the combined sequence of tokens.
It learns to generate latents in the ERP image by reasoning their relationship to latents in the perspective image in a purely data-driven way.

\heading{Generating canonical panoramas.}
Since we do not provide explicit camera pose to the model, the generated panorama $\outpano$ can be in any coordinate system.
Prior works~\cite{CubeDiff, ViewPointPanoVideo} assume the conditioning view $\inpers$ is always at the center of the ERP, hence generating panoramas with ``unnatural'' gravity directions (i.e., not pointing towards the bottom of the panorama).
However, this requires the model to learn different spherical distortion patterns depending on the actual pose of the input $\inpers$.
Our ablations show that this leads to degraded visual quality (see \cref{tab:ablate-video-canonicalization}).
Instead, we enforce a \emph{Canonical Coordinate} constraint, for which the model is trained to generate panoramas in a standard, gravity-aligned upright orientation, regardless of the camera pose of the input $\inpers$.
This requires the model to infer the camera pose of $\inpers$ to ``place'' it on a canonical 360$^\circ\!$ canvas, and generate the rest of the panorama accordingly.

Implementing the canonical training objective requires ground-truth panoramas to be consistently aligned.
This condition is naturally satisfied by our image datasets as they are predominantly synthetic renderings of 3D scenes~\cite{Structured3DDataset}.
Yet, our real-world video datasets~\cite{360-1M} frequently exhibit arbitrary \textit{non-canonical} orientations.
Thus, we design a two-stage data pre-processing pipeline.
We first apply COLMAP~\cite{Colmap} to estimate per-frame camera pose, and rotate each frame to have zero rotation relative to the first frame.
Then, we run GeoCalib~\cite{GeoCalib} to estimate the global gravity direction of the stabilized video, and rotate the video to align the gravity direction with the vertical axis.
This data pre-processing step ensures the model trains on consistent, gravity-aligned data, thereby generating canonical videos at test time.
See \Cref{app-fig:video-canonicalization} in the Appendix for an example.

\begin{figure*}[t]
  \vspace{\pagetopmargin}
  \centering
  \begin{subfigure}[b]{\textwidth}
    \centering
    \includegraphics[width=\textwidth]{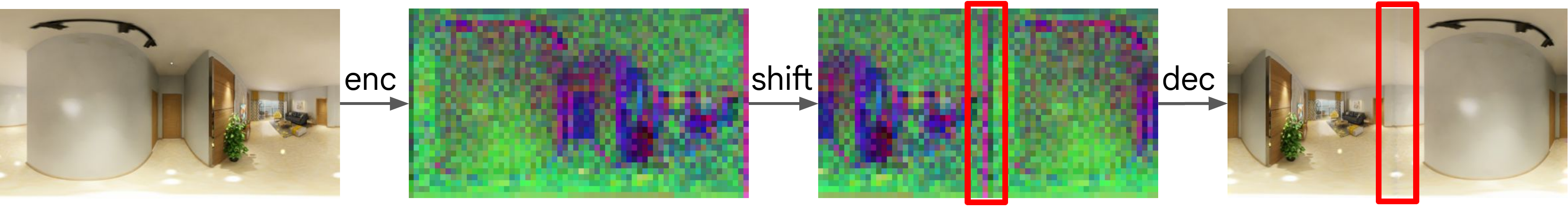}
    \vspace{-4mm}
    \caption{Naive VAE encoding of the panorama image leads to boundary discontinuities in the latent space. 
    }
  \label{fig:circluar-encoding-a}
  \end{subfigure}
  \\
  \vspace{2mm}
  \begin{subfigure}[b]{\textwidth}
    \centering
    \includegraphics[width=\textwidth]{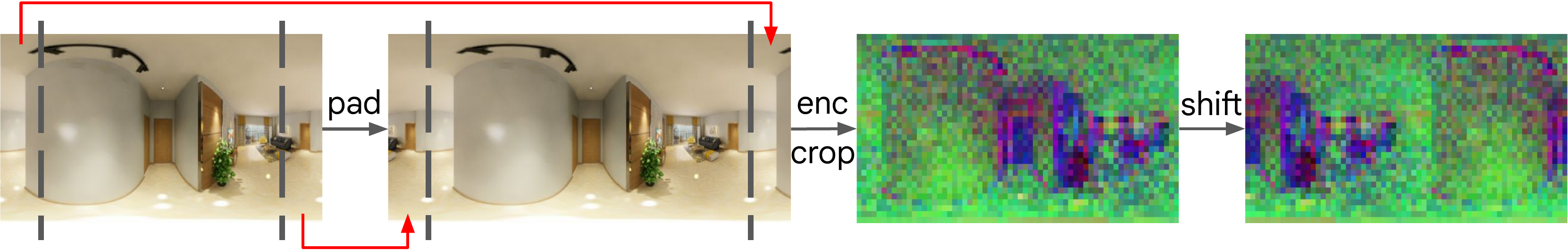}
    \caption{Our Circular Latent Encoding leads to seam-free latent representations.}
  \label{fig:circluar-encoding-b}
  \end{subfigure}
  \vspace{\figcapmargin}
  \caption{
  \textbf{Illustration of Circular Latent Encoding.}
  The top row (a) shows the seam artifact from naive VAE encoding.
  Shifting the encoded panorama latent by 180$^\circ\!$ shows a sharp discontinuity at the center, resulting in gray line-like artifacts when decoded back to image.
  The bottom row (b) illustrates our solution.
  Before encoding, we apply circular padding to the panorama image.
  After encoding, the latents in the padded regions are dropped.
  The shifted latent is now free from discontinuity, providing a seamless latent representation for diffusion training.
  }
  \label{fig:circluar-encoding}
  \vspace{\figmargin}
\end{figure*}

\subsection{Seam-free Generation via Circular Latent Encoding}
\label{subsec:seamless-pano}

A common issue in panorama generation is ``seam artifacts'', where the left and right boundaries of the ERP image have visible discontinuities when concatenated (see \cref{fig:ablate-seam-qualitative}).
Prior works often attribute this to the generation process, employing \textit{inference time} tricks such as rotated denoising (shifting the panorama cyclically across sampling steps)~\cite{PanoWan, PanoDiffusion} and circular padding in the VAE decoder~\cite{360DVDVideo}.

We argue that the root cause of seams lies not in the inference stage, but in the \emph{training stage}.
Modern diffusion models are often applied in the latent space of a convolution-based VAE.
When encoding a panorama image in the ERP format, the convolution layers perform zero-padding at the image boundaries, which introduces boundary artifacts in the feature maps~\cite{CNNPosInfo}.
As a result, even if the panorama image $\outpano$ is free from seams in pixel space, its latent representation $\outpanotoken$ contains a discontinuity (\cref{fig:circluar-encoding-a}).
We posit that this discontinuity is the root cause of seam artifacts in the generated panorama.

To eliminate the seam discontinuity in the encoded latent, we propose \emph{Circular Latent Encoding}.
Before encoding the latent of a panorama data, we crop $w'$ (set as $W/8$ in our experiments) columns from its left and right regions, and pad them to the opposite side of the panorama to extend the boundary at both sides:
\begin{equation}
    \mathrm{y_{equi}^{pad}} = \latentencoder(\mathrm{Concat}([\outpano[\mathrm{-}w'\mathrm{:}], \outpano, \outpano[\mathrm{:}w']])) .
\end{equation}
After encoding, we drop the latent corresponding to the padded regions.
This ensures the input sequence length to DiT is unchanged, thus introducing no overhead to training and inference.
This simple technique produces a seam-free latent space (\cref{fig:circluar-encoding-b}), which serves as the correct target for model training.

\section{Experiments}
\label{sec:experiments}

In this section, we conduct extensive experiments to answer the following questions:
\begin{enumerate*}[label=(\roman*)] 
\item How well does \MethodName perform on perspective-to-360$^\circ\!$ image and video generation? (\cref{subsec:pano-image-gen} and \cref{subsec:pano-video-gen})
\item How accurate are the camera FoV and orientation inferred by our model? (\cref{subsec:camera-est})
\item What is the impact of each design choice in our framework? (\cref{subsec:ablation})
\end{enumerate*}

\begin{table*}[!t]
\centering
\scriptsize
\renewcommand{\arraystretch}{1.1}
\caption{
\textbf{Quantitative results of perspective-to-360$^\circ\!$ image generation on Laval Indoor and SUN360 datasets.}
We borrow baseline results from CubeDiff \cite{CubeDiff} and report CubeDiff results under the single text description setting for a fair comparison.
\MethodName achieves a clear improvement across all metrics, only marginally lagging behind CubeDiff in terms of CLIP-FID on Laval Indoor.
}
\label{tab:imageeval}
\vspace{\tablecapmargin}
\resizebox{1.0\linewidth}{!}{
    \begin{tabular}{lcccccccccc}
        \toprule
        \multirow{2}{*}[-.2em]{\textbf{Method}} & \multicolumn{5}{c}{\textbf{Laval Indoor}} & \multicolumn{5}{c}{\textbf{SUN360}} \\
         \cmidrule(lr){2-6}\cmidrule(lr){7-11}
        & FID\,$\downarrow$ & KID ($\times 10^2$)\,$\downarrow$ & CLIP-FID\,$\downarrow$ & FAED\,$\downarrow$ & CS\,$\uparrow$ & FID\,$\downarrow$  & KID ($\times 10^2$)\,$\downarrow$  & CLIP-FID\,$\downarrow$ & FAED\,$\downarrow$ & CS\,$\uparrow$ \\
        \hline
        OmniDreamer~\cite{OmniDreamer} & 71.0 & 5.17 & 23.9 & 19.2& - & 92.3 & 8.89 & 51.7 & 30.4 & - \\
        PanoDiffusion~\cite{PanoDiffusion} & 58.6 & 4.08 & 26.6 & 106.8 & - & 52.9 & 3.51 & 28.9 & 98.0 & - \\
        Diffusion360~\cite{Diffusion360} & 33.1 & 2.07 & 16.9 & 23.7 & 26.38 & 45.4 & 3.73 & 18.5 & 12.6 & 22.89 \\
        CubeDiff~\cite{CubeDiff} & \underline{9.5} & \underline{0.32} & \textbf{3.2} & \underline{18.4} & \underline{27.02} & \underline{25.5} & \underline{1.33} & 8.1 & \underline{7.6} & 25.00 \\
        \textbf{\MethodName~(ours)} & \textbf{8.0} & \textbf{0.22} & \underline{4.6} & \textbf{9.8} & \textbf{29.21} & \textbf{22.4} & \textbf{1.27} & \textbf{7.3} & \textbf{3.8} & \textbf{28.07} \\
        \bottomrule
    \end{tabular}
    }
\vspace{\tablemargin}
\end{table*}

\begin{figure*}[t]
  \vspace{\pagetopmargin}
  \centering
    \includegraphics[width=\textwidth]{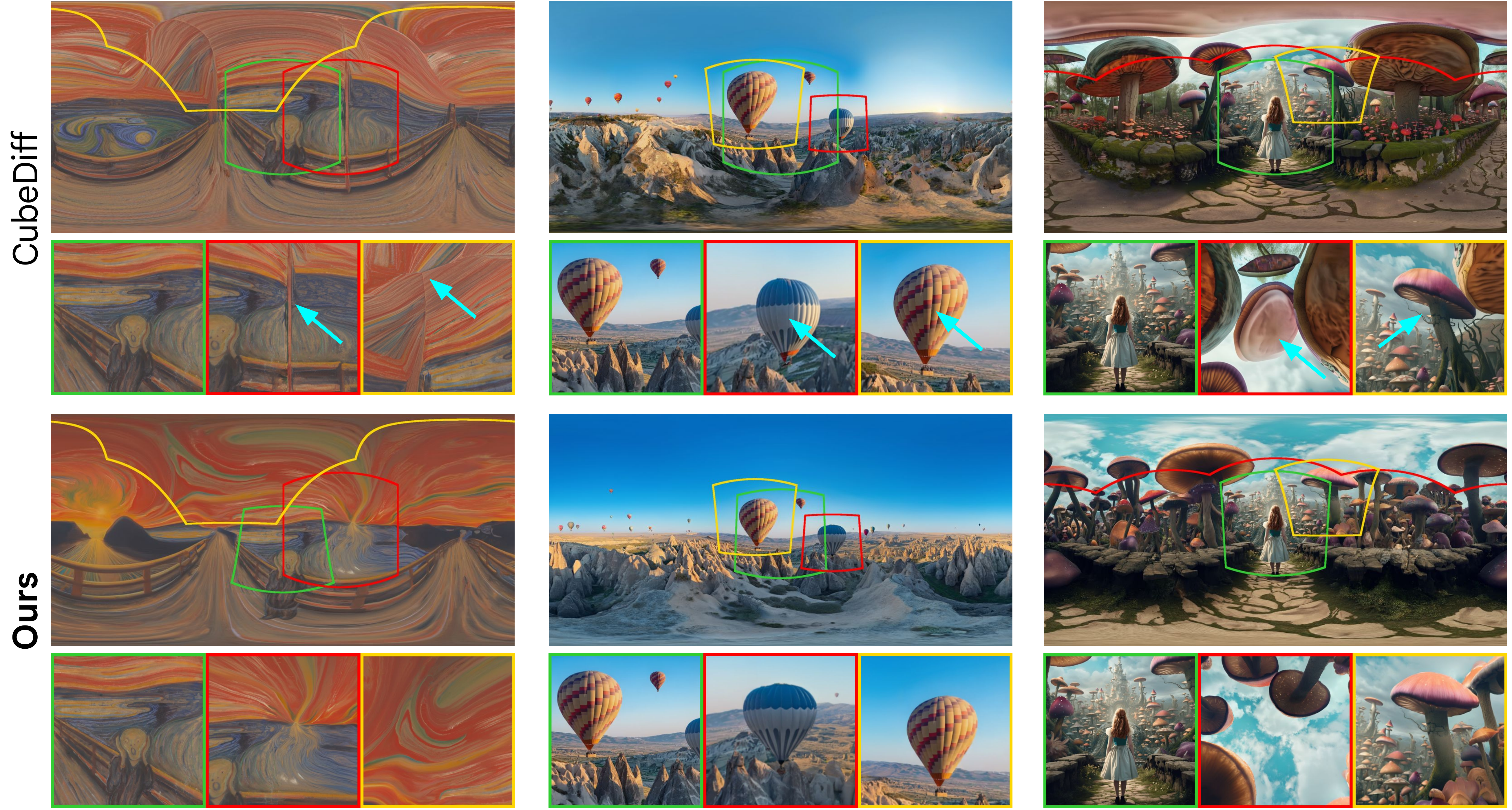}
    \vspace{\figcapmargin}
    \caption{
    \textbf{Qualitative results of perspective-to-360$^\circ\!$ image generation.}
    We show multiple perspective views projected from the panorama, where the image with the {\color{BrightGreen} \textbf{green border}} is the conditioning image.
    Due to the use of a cubemap representation, CubeDiff sometimes generates seams between faces (left).
    In addition, CubeDiff always assumes the input image has a 90$^\circ\!$ FoV; yet when the actual FoV is smaller, it has to stretch the objects at the image boundary.
    This leads to distorted object structure, e.g., the balloons (middle) and the mushroom (right).
    In contrast, \MethodName estimates the correct camera FoV and orientation of the input as shown by the green box on the panorama image, and thus produces much less distorted objects.
    Please check out our \href{https://360anything.github.io/index.html\#360-image-result}{project page} to view the generated panorama images interactively.
    }
  \label{fig:qualitative-results-image}
  \vspace{\figmargin}
\end{figure*}

\subsection{Panoramic Image Outpainting}
\label{subsec:pano-image-gen}

\heading{Implementation details.}
We fine-tune FLUX.1-dev~\cite{FLUX}, a state-of-the-art (SoTA) open weights text-to-image DiT.
We use the Adam optimizer~\cite{Adam} with a learning rate of $5\times\text{10}^{-\text{5}}$ and train with a batch size of 512 for 50k steps.
At inference time, we use FLUX's default sampler~\cite{RFSampler} with 50 sampling steps and timestep shifting of 3.16.
See Appendix~\ref{app-subsec:our-implementation} for more implementation details.

\heading{Training data and augmentations.}
For a fair comparison against CubeDiff~\cite{CubeDiff}, the prior state-of-the-art, we train on the same datasets with captions from Gemini 2.5 Flash~\cite{Gemini2_5}.
To handle input images with diverse camera setup at test time, we uniformly sample FoV in [30$^\circ$, 120$^\circ$], pitch in [$-$60$^\circ$, 60$^\circ$], roll in [$-$15$^\circ$, 15$^\circ$], and use them to crop the conditioning perspective images for training.
Following prior works, we also perform horizontal roll augmentation on the panorama image.
We train the model to generate ERP images at 1024$\times$2048 resolution.

\heading{Evaluation data and metrics.}
We follow the evaluation protocol proposed in CubeDiff and report results on the Laval Indoor~\cite{LavalIndoorDataset} and SUN360~\cite{SUN360Dataset} datasets.
To measure visual quality, we report Fréchet Inception Distance (FID)~\cite{FID}, Kernel Inception Distance (KID)~\cite{KID}, FID on CLIP~\cite{CLIP} features (CLIP-FID), and FID on features of an auto-encoder fine-tuned on panorama images (FAED)~\cite{PanFusion}.
Following prior works, FID, KID, CLIP-FID are computed on perspective crops from the generated ERP image, while FAED is computed directly on the generated ERP image.
We also report CLIP-score (CS)~\cite{CLIPScore} for text alignment.

\heading{Quantitative results.}
\Cref{tab:imageeval} compares \MethodName with several perspective-to-panorama image generation baselines.
In terms of visual quality, \MethodName substantially outperforms all baselines on both datasets across FID, KID, and FAED metrics.
Although it marginally lags behind CubeDiff in CLIP-FID on Laval Indoor ($4.6$ vs. $3.2$), it outperforms it on SUN360 ($7.3$ vs. $8.1$) which has more complex scene layouts and textures.
Notably, we achieve a significant improvement in FAED, reducing the error by nearly 50\% compared to the state-of-the-art.
FAED is the only metric evaluated on the entire panorama, and it shows that \MethodName generates 360$^\circ\!$ images with clearly better quality and geometry.
Finally, our method also achieves the best CLIP-score, demonstrating its superior capability in adhering to prompts.

\heading{Qualitative results.}
We only compare with CubeDiff since other baselines lag behind by a \textit{large} margin in all metrics.
As CubeDiff is not open-sourced, we compare with samples from their website.
CubeDiff leverages a cubemap representation with six faces, each with 90$^\circ\!$ FoV.
It treats the conditioning image as the front face and denoises the other five faces.
As shown in \Cref{fig:qualitative-results-image} left, it sometimes generates \textit{visible seams} between faces.
In addition, when the input perspective image has an FoV smaller than 90$^\circ\!$, CubeDiff has to stretch the object at the boundary, leading to distorted air-balloons (middle) and mushrooms (right).
In contrast, \MethodName accurately infers the camera parameters to place the input image in the ERP space, and produces objects with the correct structure.

\subsection{Panoramic Video Outpainting}
\label{subsec:pano-video-gen}

\heading{Implementation details.}
We fine-tune Wan2.1-14B~\cite{Wan2.1}, a SoTA open source text-to-video DiT.
We use the Adam optimizer with a learning rate of $1\times\text{10}^{-\text{5}}$ and train with a batch size of 64 for 20k steps.
For model inference, we run 50 sampling steps with Wan's default sampler and a timestep shifting of 3.0.

\heading{Training data and augmentations.}
We take the same training data from Argus~\cite{Argus}, run our video canonicalization pipeline, and caption using Gemini 2.5 Flash.
We follow Argus to simulate camera trajectories to crop conditioning perspective videos in training.
However, we note that simulated camera movement lacks diversity.
Thus, we also incorporate camera trajectories extracted from real-world videos~\cite{DynPose100K}, which improves generalization to in-the-wild videos (see \cref{app-fig:ablate-real-world-cam}
in the Appendix).
We train on 512$\times$1024 resolution videos with 81 frames.

\heading{Evaluation data and metrics.}
We adopt the 101 testing videos from Argus~\cite{Argus}.
The conditioning perspective videos come from two types of camera trajectories, namely, simulated and extracted from real-world videos.
To measure input preservation, we report PSNR and LPIPS~\cite{LPIPS} between ground-truth and generated panorama videos within regions covered by the perspective video.
We also report FVD~\cite{FVD}, Imaging Quality, Aesthetic Quality, and Motion Smoothness from VBench~\cite{VBench} to evaluate overall quality.
Note that VBench metrics are computed on \textit{perspective} crops (left, right, front, back) of the generated panorama videos.

\begin{table*}[!t]
\centering
\scriptsize
\renewcommand{\arraystretch}{1.1}
\caption{
\textbf{Quantitative results of perspective-to-360$^\circ\!$ video generation.}
We follow Argus~\cite{Argus} to evaluate on two sets of camera trajectories.
\emph{Imag.}, \emph{Aes.}, and \emph{Motion} stand for the Imaging Quality, Aesthetic Quality, and Motion Smoothness metrics from VBench~\cite{VBench}.
Since the exact eval split used in Argus is unavailable, we reproduce the eval split based on communication with the author (* denotes results on it).
We report \textit{both} the original and our reproduced results for Argus, which closely match to validate our reproduced eval set.
Our method outperforms all baselines across all metrics.
}
\label{tab:videoeval}
\vspace{\tablecapmargin}
\resizebox{1.0\linewidth}{!}{
    \begin{tabular}{lcccccccccccc}
        \toprule
         \multirow{2}{*}[-.2em]{\textbf{Method}} & \multicolumn{6}{c}{\textbf{Real camera trajectory}} & \multicolumn{6}{c}{\textbf{Simulated camera trajectory}} \\
         \cmidrule(lr){2-7}\cmidrule(lr){8-13}
         & PSNR\,$\uparrow$ & LPIPS\,$\downarrow$ & FVD\,$\downarrow$ & Imag.\,$\uparrow$ & Aes.\,$\uparrow$ & Motion\,$\uparrow$ & PSNR\,$\uparrow$ & LPIPS\,$\downarrow$ & FVD\,$\downarrow$ & Imag.\,$\uparrow$ & Aes.\,$\uparrow$ & Motion\,$\uparrow$ \\
        \midrule
        Imagine360*~\cite{Imagine360} & 21.00 & 0.2636 & 1398.9 & 0.4556 & 0.4658 & 0.9666 & 20.45 & 0.2719 & 1532.1 & 0.4485 & 0.4536 & 0.9725 \\
        Argus~\cite{Argus} & 21.83 & 0.2409 & 1228.6 & 0.4939 & 0.4828 & 0.9802 & 21.50 & 0.2602 & 1100.1 & 0.4812 & 0.4784 & 0.9805 \\
        Argus*~\cite{Argus} & 22.35 & 0.2310 & 1020.7 & 0.4971 & 0.4863 & 0.9728 & 21.10 & 0.2653 & 1127.2 & 0.4762 & 0.4682 & \underline{0.9852} \\
        ViewPoint*~\cite{ViewPointPanoVideo} & \underline{23.25} & \underline{0.1364} & \underline{844.3} & \underline{0.5293} & \underline{0.5150} & \underline{0.9881} & \underline{22.77} & \underline{0.1326} & \underline{957.8} & \underline{0.5105} & \underline{0.5045} & 0.9827 \\
        \textbf{\MethodName~(ours)}& \textbf{25.75} & \textbf{0.0468} & \textbf{483.4} & \textbf{0.5515} & \textbf{0.5427} & \textbf{0.9885} & \textbf{23.64} & \textbf{0.0846} & \textbf{432.9} & \textbf{0.5489} & \textbf{0.5394} & \textbf{0.9891} \\
        \bottomrule
    \end{tabular}
    }
\vspace{\tablemargin}
\end{table*}

\begin{figure*}[!t]
  \vspace{\pagetopmargin}
  \centering
    \includegraphics[width=\textwidth]{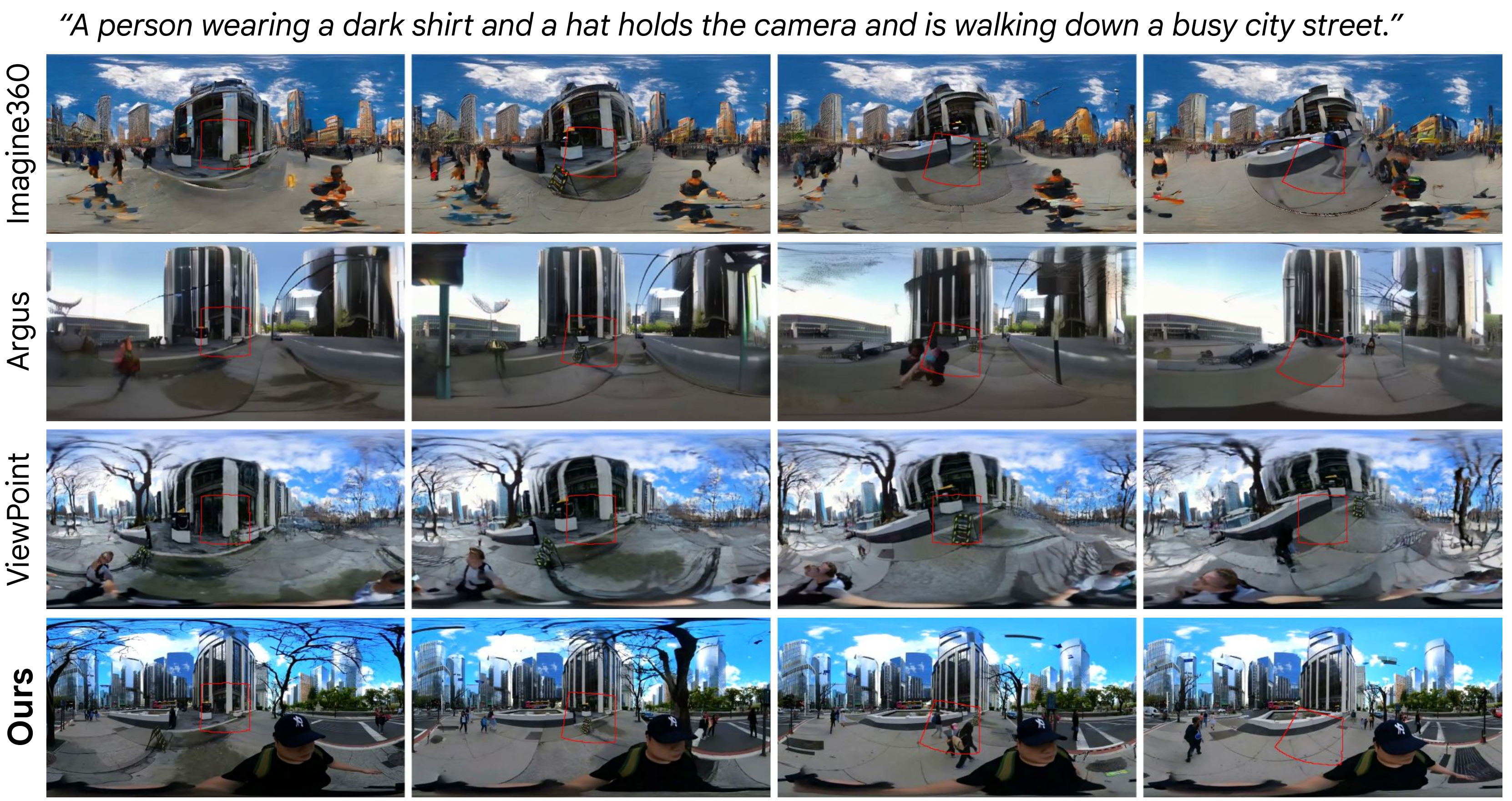}
    \vspace{\figcapmargin}
    \caption{
    \textbf{Qualitative results of perspective-to-360$^\circ\!$ video generation.}
    Regions corresponding to the input conditioning video are highlighted in {\color{red} \textbf{red}}.
    Both Imagine360 and Argus exhibit low visual quality and distortions.
    ViewPoint always places the conditioning video at the center of the output, and thus generates a rotated image when the video contains large camera motion, leading to distortions~(e.g., people and buildings).
    In contrast, \MethodName generates stably \textit{canonicalized} panorama videos, and accurately follows the text prompt to outpaint a person holding the camera.
    Please see our \href{https://360anything.github.io/index.html\#compare-with-baseline}{project page} for better visual comparisons in the video format.
    }
  \label{fig:qualitative-results-video}
  \vspace{\figmargin}
\end{figure*}

\heading{Quantitative results.}
\Cref{tab:videoeval} compares \MethodName with recent perspective-to-360$^\circ\!$ video generation methods.
We outperform all baselines across all metrics on both subsets, often by a large margin.
Surprisingly, while prior works construct ``pixel-aligned'' input by unprojecting the conditioning perspective video, our method achieves better PSNR and LPIPS, meaning that it learns to better preserve the conditioning perspective video in the output.
The significantly lower FVD indicates that our generated panorama videos exhibit more natural spherical distortion in the ERP format.
In addition, \MethodName achieves higher VBench scores, demonstrating its superior visual and motion quality.

\heading{Qualitative results.}
As shown in \Cref{fig:qualitative-results-video}, \MethodName exhibits significantly higher visual quality than Imagine360~\cite{Imagine360} and Argus~\cite{Argus}.
Since ViewPoint~\cite{ViewPointPanoVideo} assumes the input video is always at the center of the ERP, it has to rotate the generated panorama given a tilted perspective video, leading to distortions~(e.g., people and buildings).
In contrast, our method generates canonicalized panorama frames that are \textit{both} temporally consistent and distortion-free.

\begin{figure*}[!t]
  \vspace{\pagetopmargin}
  \centering
    \includegraphics[width=\textwidth]{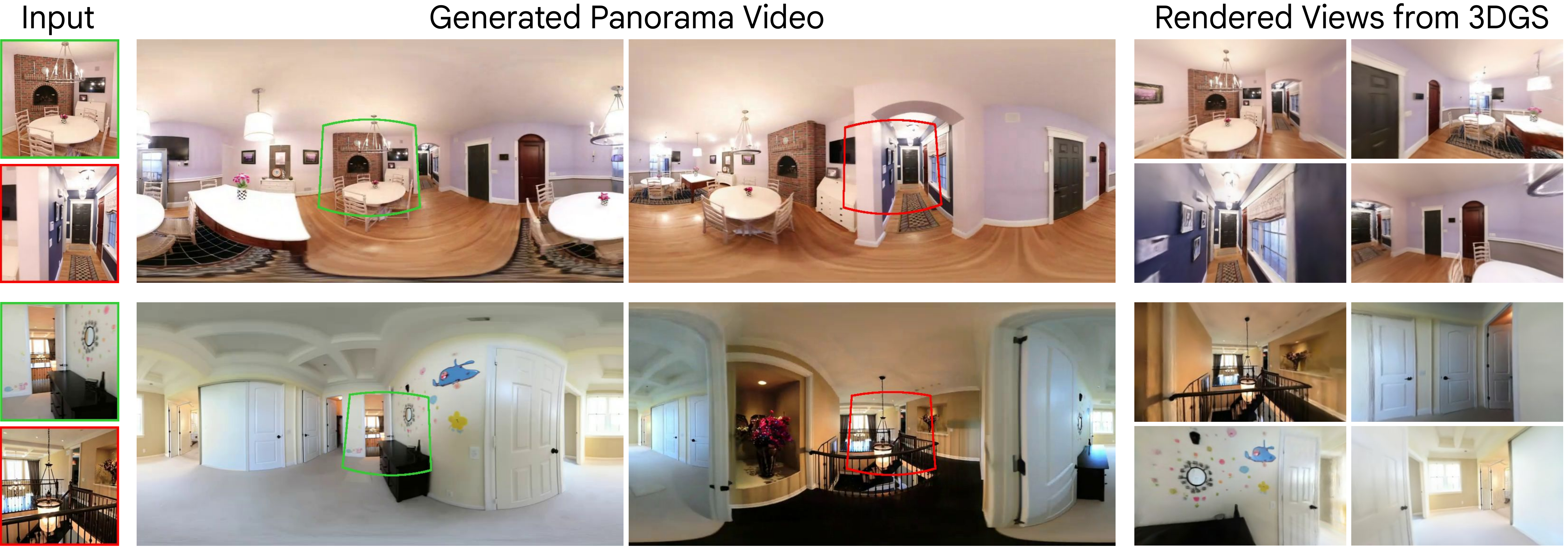}
    \vspace{\figcapmargin}
    \caption{
    \textbf{Qualitative results of 3D scene reconstruction.}
    Given an input monocular video (left), \MethodName outpaints the whole 360$^\circ\!$ viewpoint (middle), from which we can optimize a 3DGS (right).
    This allows fly-through exploration of the entire 3D scene.
    Please check out our \href{https://360anything.github.io/index.html\#3d-scene-recon}{project page} to view 360$^\circ\!$ rendering of the 3DGS.
    }
  \label{fig:3dgs-qualitative}
  \vspace{\figmargin}
\end{figure*}

\heading{3D scene reconstruction.}
To demonstrate the 3D consistency of our generated videos, we distill the panoramic videos produced by \MethodName into a 3D Gaussian Splat (3DGS)~\cite{3DGS}.
We only test videos of static scenes such as indoor rooms~\cite{RealEstate10k, ScanNetPP} as vanilla 3DGS cannot handle dynamic subjects.
This process involves two main steps:
\begin{enumerate*}[label=(\roman*)] 
\item First, we employ rig-based COLMAP~\cite{ColmapRigs, Colmap} on the generated panoramic video to obtain camera poses.
We project each video frame to cubemap faces and perform constrained bundle adjustment using a cubemap rig.
\item Then, we train a vanilla 3DGS on the posed images.
The scene reconstruction results are qualitatively shown in \Cref{fig:3dgs-qualitative}.
\end{enumerate*}
Given a monocular video with partial scene coverage, \MethodName outpaints the full 360$^\circ\!$ view that provides enough geometric cues for 3D reconstruction.
This allows free exploration over the reconstructed scene, demonstrating its strong geometry consistency.

\subsection{Single Image Camera Calibration}
\label{subsec:camera-est}

\heading{Experimental setup.}
To measure the camera parameters estimated by our model, we evaluate \MethodName (the image version) on several benchmarks on FoV and camera pose estimation.
Given a testing image $\inpers$, we generate a panorama image $\outpano$, and infer the camera metadata via exhaustive search:
\begin{equation}
    \min_{\mathrm{fov, pitch, roll}}\ \| \, \inpers - \mathrm{pano2pers}(\outpano; \mathrm{fov, pitch, roll}) \, \|^2 \, , \nonumber
\end{equation}
where $\mathrm{pano2pers}()$ is the panorama-to-perspective image projection function, and~$\mathrm{(fov, pitch, roll)}$ are the estimated camera metadata.

\begin{table*}[!t]
    \centering
    \begin{minipage}[t]{0.56\linewidth}
        \caption{
            \textbf{Camera FoV estimation results (in degrees).}
            We borrow baseline numbers from the MoGe paper~\cite{MoGe}.
            Our zero-shot FoV estimation approach outperforms several supervised baselines, and lagging only slightly behind state-of-the-art methods DUSt3R and MoGe.
        }
        \vspace{\tablecapmargin}
        \label{tab:foveval}
        \centering
        \scriptsize
        \setlength{\tabcolsep}{1pt}
        \renewcommand{\arraystretch}{1.1}
        \resizebox{1.0\linewidth}{!}{
        \begin{tabular}{lcccccc}
            \toprule
            \multirow{2}{*}[-.3em]{\textbf{Method}} & \multicolumn{2}{c}{\textbf{NYUv2}} & \multicolumn{2}{c}{\textbf{ETH3D}} & \multicolumn{2}{c}{\textbf{iBims-1}} \\
            \cmidrule(lr){2-3}
            \cmidrule(lr){4-5}
            \cmidrule(lr){6-7}
            & Mean\,$\downarrow$ & Med.\,$\downarrow$ & Mean\,$\downarrow$ & Med.\,$\downarrow$ & Mean\,$\downarrow$ & Med.\,$\downarrow$ \\
            \midrule
            Perspective~\cite{jin2022PerspectiveFields} & 5.38 & 4.39 & 13.6 & 11.9 & 10.6 & 9.30 \\
            WildCam~\cite{WildCam} & 3.82 & 3.20 & 7.70 & 5.81 & 9.48 & 9.08 \\
            LeReS~\cite{LeReS} & 19.4 & 19.6 & 8.26 & 7.19 & 18.4 & 17.5 \\
            UniDepth~\cite{UniDepth} & 7.56 & 4.31 & 10.7 & 9.96 & 11.9 & 5.96 \\
            DUSt3R~\cite{DUSt3R} & \textbf{2.57} & \textbf{1.86} & 5.77 & \underline{3.60} & \underline{3.83} & \underline{2.53} \\
            MoGe~\cite{MoGe} & \underline{3.41} & 3.21 & \textbf{2.50} & \textbf{1.54} & \textbf{2.81} & \textbf{1.89} \\
            \midrule
            \textbf{\MethodName} & 3.90 & \underline{3.17} & \underline{5.68} & 4.22 & 5.21 & 4.04 \\
            \bottomrule
        \end{tabular}
        }
    \end{minipage}
    \hfill
    \begin{minipage}[t]{0.41\linewidth}
        \caption{
            \textbf{Camera pose estimation results (in degrees).}
            Baseline numbers are from GeoCalib~\cite{GeoCalib}.
            Our zero-shot approach outperforms most supervised baselines and lags behind current state-of-the-art, GeoCalib, by just $\sim$0.5 degrees.
        }
        \vspace{\tablecapmargin}
        \label{tab:cameraposeeval}
        \centering
        \scriptsize
        \setlength{\tabcolsep}{1pt}
        \renewcommand{\arraystretch}{1.1}
        \resizebox{1.0\linewidth}{!}{
        \begin{tabular}{lcccc}
            \toprule
            \multirow{2}{*}[-.3em]{\textbf{Method}} & \multicolumn{2}{c}{\textbf{MegaDepth}} & \multicolumn{2}{c}{\textbf{LaMAR}} \\
            \cmidrule(lr){2-3}
            \cmidrule(lr){4-5}
            & Roll\,$\downarrow$ & Pitch\,$\downarrow$ & Roll\,$\downarrow$ & Pitch\,$\downarrow$ \\
            \midrule
            MSCC~\cite{Song2024MSCC} & 0.90 & 5.73 & 1.44 & 3.02 \\
            ParamNet~\cite{jin2022PerspectiveFields} & 1.17 & 3.99 & 0.93 & 2.15 \\
            UVP~\cite{Pautrat_2023_UncalibratedVP} & \underline{0.51} & 4.59 & \underline{0.38} & 1.34 \\
            GeoCalib~\cite{GeoCalib} & \textbf{0.36} & \textbf{1.94} & \textbf{0.28} & \textbf{0.87} \\
            \midrule
            \textbf{\MethodName} & 0.87 & \underline{2.56} & 0.68 & \underline{1.23} \\
            \bottomrule
        \end{tabular}
        }
    \end{minipage}
    \vspace{\tablemargin}
\end{table*}

\heading{Results on FoV estimation.}
We follow MoGe~\cite{MoGe} and test our model on three real-world datasets: NYUv2~\cite{NYUv2Dataset}, ETH3D~\cite{ETH3DDataset}, and iBims-1~\cite{iBims1Dataset}.
\Cref{tab:foveval} compares \MethodName with several baselines.
It is worth mentioning that all baseline models are trained on large-scale datasets \textit{specifically} for 3D understanding tasks, while \MethodName is only trained for image-to-360$^\circ\!$ outpainting.
In addition, over 90\% of our training images are indoor scenes~\cite{Structured3DDataset}, thus creating a large \textit{domain gap} with the outdoor ETH3D and iBims-1 datasets.
Nevertheless, our zero-shot FoV estimation approach ranks among top-3 on most of the datasets.
It achieves a low average estimation error of only 4.93 degrees, outperforming several supervised baselines while only lagging slightly (by $1-2$ degrees) behind recent methods DUSt3R~\cite{DUSt3R} (4.06) and MoGe (2.91).

\heading{Results on camera orientation estimation.}
We follow GeoCalib~\cite{GeoCalib} to test our model on MegaDepth~\cite{MegaDepthDataset} and LaMAR~\cite{LaMARDataset} datasets for estimating camera roll and pitch angles from a single image.
As shown in \Cref{tab:cameraposeeval}, our zero-shot approach clearly outperforms most \textit{supervised} baselines, only lagging behind the current state-of-the-art GeoCalib~($\sim$0.5 degrees in both roll and pitch).
These results demonstrate that our method learns to establish accurate correspondence between the conditioning perspective input and the generated panorama image.

\begin{table*}[!t]
\caption{
\textbf{Comparison of seam elimination techniques in perspective-to-360$^\circ\!$ image and video tasks.}
We report the discontinuity score (DS) to measure seams at the boundary of the panorama.
\textit{CLE} stands for our Circular Latent Encoding technique.
}
\vspace{\tablecapmargin}
\label{tab:ablate-seam}
\centering
\scriptsize
\setlength{\tabcolsep}{3.5pt} %
\renewcommand{\arraystretch}{1.1}
\begin{tabular}{lcccccc}
    \toprule
    \multirow{2}{*}[-.3em]{\textbf{Method}} & \multicolumn{3}{c}{\textbf{Image}} & \multicolumn{3}{c}{\textbf{Video}} \\
    \cmidrule(lr){2-4}\cmidrule(lr){5-7}
    & Vanilla & Blended Decoding & \textbf{CLE~(ours)} & Vanilla & Blended Decoding & \textbf{CLE~(ours)} \\
    \hline
    DS\,$\downarrow$ & 9.92 & 5.29 & \textbf{3.87} & 35.52 & 19.84 & \textbf{13.28} \\
    \bottomrule
\end{tabular}
\vspace{\tablemargin}
\end{table*}

\begin{figure*}[!t]
  \centering
    \includegraphics[width=\textwidth]{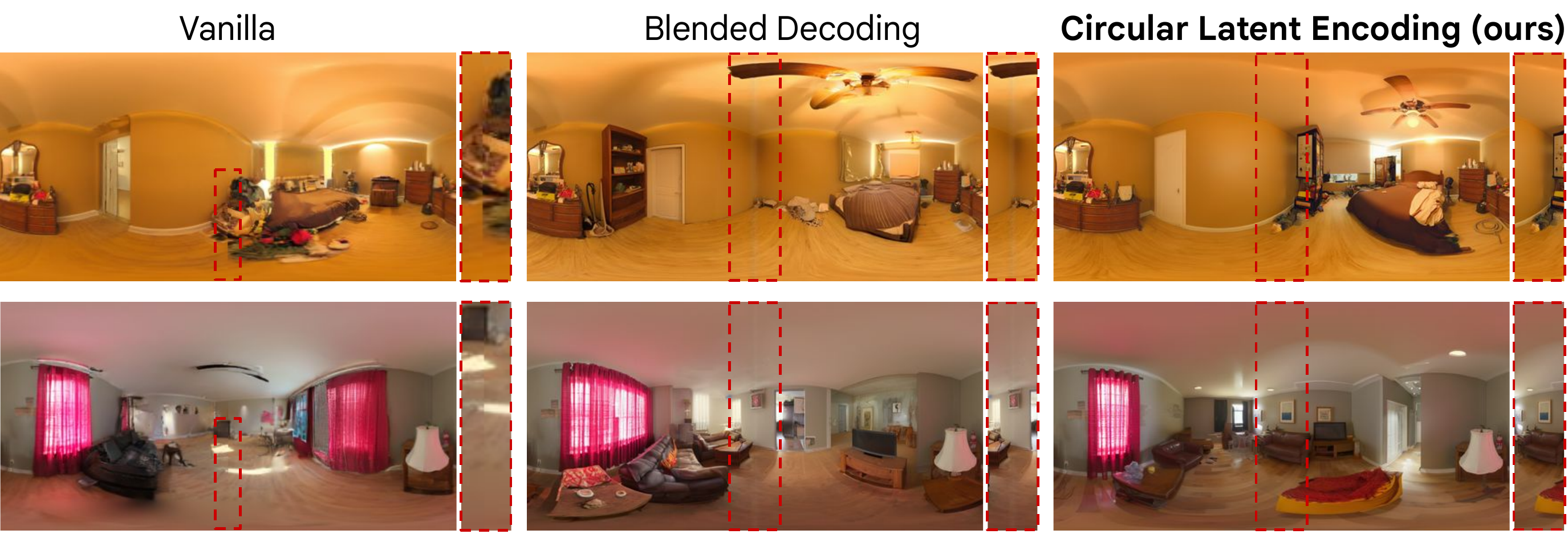}
    \vspace{\figcapmargin}
    \caption{
    \textbf{Qualitative evaluation of various seam elimination techniques.}
    For ease of visualization, we shift the generated panorama by 180$^\circ\!$ to show the concatenation of its left and right boundaries.
    Without any intervention (left), there are clear seams.
    Blended Decoding~\cite{Argus} (middle) ``blurs'' the seam to remove discontinuities; yet, it introduces gray line-like artifacts.
    Our technique (right) eliminates boundary artifacts entirely.
    We recommend zooming-in to evaluate these differences appropriately.
   }
  \label{fig:ablate-seam-qualitative}
  \vspace{\figmargin}
\end{figure*}

\begin{table*}[t]
\vspace{1mm}
\caption{
\textbf{Robustness to input view variations.} 
We report the absolute FID/FAED for the standard view $(90^\circ, 0^\circ, 0^\circ)$ and the relative change in metrics ($\Delta$) for other input views.
The last column shows the average degradation across all input view variations.
\textit{w/o Camera Aug.} stands for our method trained without camera augmentation, i.e., always train with $(90^\circ, 0^\circ, 0^\circ)$ as the conditioning view.
\textit{CC w/ GT Camera} denotes the channel concatenation model that has access to ground-truth camera information.
\MethodName shows similar robustness to it even without camera metadata.
}
\vspace{\tablecapmargin}
\label{tab:diff-camera-image-result}
\centering
\scriptsize
\renewcommand{\arraystretch}{1.1}
\resizebox{1.0\linewidth}{!}{
\begin{tabular}{clc|cccccc c}
    \toprule
    \multirow{2}{*}[-.2em]{\textbf{Metric}} & \multirow{2}{*}[-.2em]{\textbf{Method}} & \multicolumn{7}{c}{\textbf{Conditioning View} (FoV, Pitch, Roll)} & \multirow{2}{*}[-.2em]{\textbf{Avg.}} \\
    \cmidrule(lr){3-9}
    & & (90, 0, 0) & (30, 0, 0) & (60, 0, 0) & (90, 30, 0) & (90, -30, 0) & (90, 30, 5) & (90, -30, -5) & \\
    \midrule
    
    \multirow{3}{*}{\textbf{FID} $\downarrow$} 
    & w/o Camera Aug. & 8.4 & +2.4 & +0.8 & +8.9 & +9.2 & +8.9 & +8.7 & +6.48 \\
    & CC w/ GT Camera & 7.7 & +3.4 & +0.8 & \textbf{+0.5} & +0.9 & \textbf{+0.5} & +0.9 & +1.17 \\
    & \textbf{\MethodName~(ours)} & 8.0 & \textbf{+2.2} & +0.8 & +0.6 & +0.9 & +0.6 & \textbf{+0.8} & \textbf{+0.98} \\
    
    \midrule
    
    \multirow{3}{*}{\textbf{FAED} $\downarrow$} 
    & w/o Camera Aug. & 10.4 & +6.6 & +1.7 & +1.4 & +1.8 & +1.4 & +1.6 & +2.42 \\
    & CC w/ GT Camera & 9.9 & \textbf{+5.1} & +1.5 & +0.4 & +0.4 & +0.3 & \textbf{+0.2} & \textbf{+1.32} \\
    & \textbf{\MethodName~(ours)} & 9.8 & +5.4 & +1.5 & \textbf{+0.3} & +0.4 & +0.3 & +0.4 & +1.38 \\
    \bottomrule
\end{tabular}
}
\vspace{\tablemargin}
\end{table*}

\subsection{Ablation Study}
\label{subsec:ablation}

\heading{Circular latent encoding.}
\Cref{tab:ablate-seam} and \Cref{fig:ablate-seam-qualitative} compare different seam elimination techniques.
We report the discontinuity score~(DS)~\cite{OmniFID-DS} to quantify seam artifacts.
Our \textit{circular latent encoding}~(CLE) dramatically reduces DS compared to the blended decoding approach proposed in Argus~\cite{Argus}.
In addition, our method introduces no overhead to the generation process.

\heading{Camera augmentation.}
Training with randomly sampled camera FoVs and orientations enables \MethodName to handle arbitrary perspective input at test time.
However, the common evaluation protocol of perspective-to-360$^\circ\!$ image generation always uses conditioning images with 90$^\circ\!$ FoV and zero pitch and roll.
We thus study whether disabling random camera sampling can further improve model performance.
Surprisingly, \Cref{tab:diff-camera-image-result} shows that such camera augmentation improves results on all metrics.
We hypothesize that training with a wide distribution of camera setup forces the model to better understand perspective-equirectangular geometry, preventing overfitting to a single mapping.

\heading{Robustness to camera parameters.}
We condition \MethodName on perspective images with different camera FoVs and orientations.
As shown in~\Cref{tab:diff-camera-image-result}, all metrics improve as FoV increases, as a larger FoV gives more information about the entire panorama image.
Performance degrades slightly when changing pitch and roll angles, yet the degradations are all less than 1.0.
We further compare with a variant of \MethodName that uses channel-concatenation instead of sequence-concatenation for conditioning (dubbed \textit{CC w/ GT Camera}).
At test time, it uses \emph{ground-truth} camera metadata to perform the perspective-to-equirectangular projection.
\Cref{tab:diff-camera-image-result} shows that it suffers from a similar performance drop as \MethodName; yet our method does not rely on camera metadata.
Overall, these results demonstrate the robustness of \MethodName.

\begin{table}[!t]
\vspace{2mm}
\caption{
    \textbf{Ablation on training video canonicalization.}
    To save compute, models are trained at a lower resolution of 256$\times$512.
    \emph{Imag.}, \emph{Aes.}, and \emph{Motion} stand for Imaging Quality, Aesthetic Quality, and Motion Smoothness from VBench~\cite{VBench}.
    Training on canonical videos improves visual quality as indicated by FVD and VBench metrics.
}
\vspace{\tablecapmargin}
\centering
\scriptsize
\resizebox{1.0\linewidth}{!}{
    \begin{tabular}{lcccccccccccc}
        \toprule
         \multirow{2}{*}[-.2em]{\textbf{Canonical}} & \multicolumn{6}{c}{\textbf{Real camera trajectory}} & \multicolumn{6}{c}{\textbf{Simulated camera trajectory}} \\
         \cmidrule(lr){2-7}\cmidrule(lr){8-13}
         & PSNR\,$\uparrow$ & LPIPS\,$\downarrow$ & FVD\,$\downarrow$ & Imag.\,$\uparrow$ & Aes.\,$\uparrow$ & Motion\,$\uparrow$ & PSNR\,$\uparrow$ & LPIPS\,$\downarrow$ & FVD\,$\downarrow$ & Imag.\,$\uparrow$ & Aes.\,$\uparrow$ & Motion\,$\uparrow$ \\
        \midrule
        No & \textbf{26.56} & \textbf{0.0491} & 559.5 & 0.4689 & 0.4939 & 0.9894 & \textbf{24.66} & \textbf{0.0656} & 527.0 & 0.4601 & 0.4917 & 0.9888 \\
        \textbf{Yes~(ours)} & 24.02 & 0.0521 & \textbf{470.8} & \textbf{0.5437} & \textbf{0.5180} & \textbf{0.9899} & 22.39 & 0.0880 & \textbf{449.8} & \textbf{0.5387} & \textbf{0.5154} & \textbf{0.9903} \\
        \bottomrule
    \end{tabular}
    }
\label{tab:ablate-video-canonicalization}
\vspace{\tablemargin}
\end{table}

\heading{Training video canonicalization.}
We ablate the effect of training on canonicalized panorama videos in \Cref{tab:ablate-video-canonicalization}.
When training on non-canonicalized videos, we always place the conditioning view at the center of the output image.
This makes reconstructing the conditioning frame \textit{much} easier, leading to better PSNR and LPIPS.
However, it degrades the visual quality and fidelity significantly, as clearly indicated by FVD and VBench metrics.
This is because the model has to generate panorama frames with varying gravity directions when the input video has non-zero roll and pitch angles, forcing it to learn different spherical distortion patterns.
In contrast, we train \MethodName to generate gravity-aligned upright panoramas, which greatly simplifies the generation task.

\section{Conclusion}
\label{sec:conclusion}

We present \MethodName, a geometry-free framework for in-the-wild perspective-to-panorama generation.
By shifting from explicit geometric unprojection to simple sequence concatenation within a DiT, we eliminate the dependency on camera information, allowing the model to learn geometric correspondence purely from data.
Furthermore, we identify VAE encoder padding as the root cause of seam artifacts and introduce a novel and principled fix.
Our approach not only achieves state-of-the-art results on panoramic image and video benchmarks, but also demonstrates robust zero-shot generalization to diverse, real-world media.
We discuss the limitations and future directions in Appendix~\ref{app-sec:limitations}.

\section*{Acknowledgment}

We thank Nikolai Kalischek for help with the image data and Rundong Luo for help with the video data.
We thank Noah Snavely and Richard Tucker for implementing the pipeline for running structure-from-motion at scale.
We thank George Kopanas for help with the 3DGS experiments, Sara Sabour for discussion on 3D SfM, Lala Li for guidance on Gemini captioning, and Charles Herrmann and Jon Shlens for their feedback.
We also thank the maxdiffusion contributors for converting FLUX and Wan from PyTorch to Jax.
Some testing videos are from Veo, Genie 3, Sora, Gen4.5, and Wan; we thank them for sharing the videos.


%
%
\bibliographystyle{splncs04}
\bibliography{main}

\clearpage
\appendix
\crefalias{section}{appendix}
\crefalias{subsection}{appendix}

\section{Detailed Experimental Setup}
\label{app-sec:more-exp-setup}

In this section, we provide full details on the datasets, baselines, evaluation settings, and the training and inference implementation details of our model.

\subsection{Training Data}
\label{app-subsec:training-data}

\textbf{We provide the prompts for data captioning on our \href{https://360anything.github.io/index.html\#data-caption}{project page}}.

\heading{Image data.}
To facilitate a fair comparison with the previous state-of-the-art method, CubeDiff~\cite{CubeDiff}, we follow them to use the same panorama image datasets as training data.
These include Polyhaven~\cite{PolyHavenDataset}, Humus~\cite{HumusDataset}, Structured3D~\cite{Structured3DDataset}, and Pano360~\cite{Pano360Dataset}.
For Structured3D, we use all three subsets, namely, empty, simple, and full.
As a result, around 90\% of data are synthetic rendering of indoor rooms from Structured3D.
We then use Gemini 2.5 Flash~\cite{Gemini2_5} for captioning.

\noindent
\textit{Data augmentation}: to handle input images with diverse camera setup at test time, we uniformly sample FoV in [30$^\circ$, 120$^\circ$], pitch in [$-$60$^\circ$, 60$^\circ$], roll in [$-$15$^\circ$, 15$^\circ$], and use them to crop the conditioning perspective images for training.
We also perform horizontal roll augmentation on the panorama image.

\begin{figure*}[!t]
  \centering
    \includegraphics[width=\textwidth]{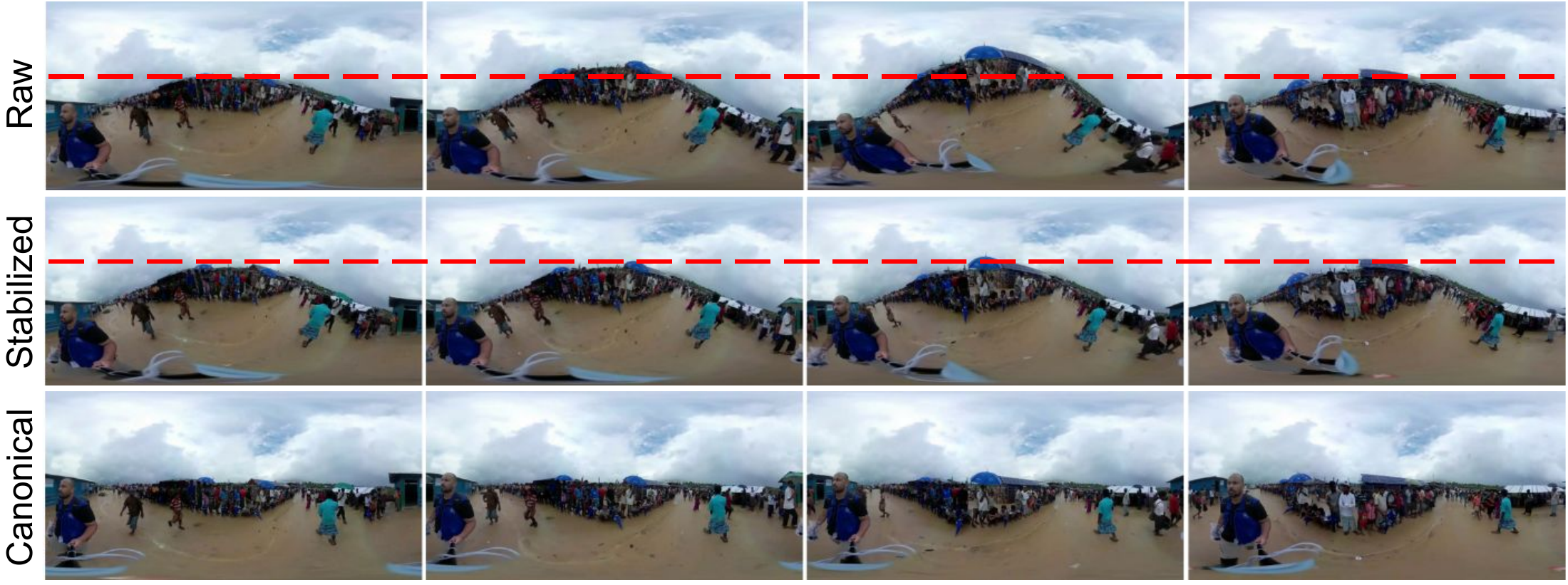}
    \vspace{\figcapmargin}
    \caption{
    \textbf{Visualization of the video canonicalization pipeline.}
    \textit{Top}: Raw panorama frames exhibit varying elevation angles, causing the horizon to fluctuate relative to the reference line ({\color{red} \textbf{red}} dashed).
    \textit{Middle}: After stabilization, inter-frame rotation is removed, resulting in a temporally consistent horizon height across all frames.
    \textit{Bottom}: After aligning the gravity direction to the vertical axis, the horizon is rectified to a straight line parallel to the image boundaries, ensuring an upright orientation.
    Please refer to our \href{https://360anything.github.io/index.html\#video-canonicalization}{project page} for better comparisons in video format.
    }
  \label{app-fig:video-canonicalization}
  \vspace{\figmargin}
\end{figure*}

\heading{Video data.}
We use the panorama videos from the 360-1M dataset~\cite{360-1M} featuring YouTube videos.
Specifically, we take the filtered subset from Argus~\cite{Argus}, which removes videos with non-panorama format, very low motion, and bad visual quality.
We then run our two-step video canonicalization pipeline (see \Cref{app-fig:video-canonicalization}).
First, we run COLMAP~\cite{Colmap} with rig support to estimate per-frame camera pose, and then rotate each frame to eliminate \emph{inter-frame camera rotation} (i.e., stabilize the video).
To improve the robustness of COLMAP, we initialize the camera pose with ORB-SLAM2~\cite{ORB-SLAM} by running on the front view cropped from the panorama.
Note that we use the modified version of ORB-SLAM2 following the changes proposed in \cite{RealEstate10k}.
Then, we run GeoCalib~\cite{GeoCalib} to convert the video to a gravity-aligned upright pose.
Since the video is stabilized, all frames should share the same gravity direction.
We thus predict the direction for all frames and then average the predictions after removing outliers (i.e., values more than 3 standard deviations from the mean).
GeoCalib is only trained on perspective images; thus we project each panorama video frame to eight perspective images (elevation=0 with uniform azimuth), run GeoCalib on each image, and take an average after the same outlier removal technique.
Finally, we rotate the video so that its gravity direction is aligned with the vertical axis.\\
Similar to the image data, we use Gemini 2.5 Flash to caption the video (downsampled to 1 FPS).
We apply the same pipeline to both the coarse and high-quality subset from Argus.

\noindent
\textit{Data augmentation}: to handle in-the-wild perspective video, we follow Argus to simulate camera trajectories with randomly sampled linear motion plus noise.
However, we note that simple linear motion is not diverse enough, and models trained on it fail to generalize to videos with complex motion (see \cref{app-fig:ablate-real-world-cam}).
Thus, we also incorporate camera trajectories extracted from real-world videos~\cite{DynPose100K, ScanNetPP} during training.
We randomly sample from simulated (80\%) and real-world (20\%) trajectories to crop perspective videos as model conditioning.

\subsection{Implementation details}
\label{app-subsec:our-implementation}

\heading{Image model.}
We fine-tune FLUX.1-dev~\cite{FLUX}, a popular open weights text-to-image diffusion transformer model.
The target panorama and conditioning perspective images are separately encoded with the VAE, flattened to 1D sequence of tokens, and concatenated along the sequence dimension as model input.
We use the same spatial index (x, y coordinate of the token) to apply 3D RoPE~\cite{RoPE} to conditioning and target tokens.
To distinguish between them in the concatenated sequence, we offset the time dimension index by $1$ when applying 3D RoPE to perspective tokens following~\cite{FLUX-Kontext, Qwen-Image}.
We fine-tune the entire model using the Adam optimizer~\cite{Adam} with a batch size of 512 for 50k steps.
The learning rate linearly increases from $0$ to $5\times\text{10}^{-\text{5}}$ in the first 1k steps, and then stays constant.
A gradient clipping of 1.0 is applied to stabilize training.
To apply classifier-free guidance (CFG)~\cite{CFG}, we randomly drop the text embedding of the caption and the conditioning image with a 10\% probability during training.
The model is trained on panorama images in the Equirectangular Projection (ERP) format with a resolution of 1024$\times$2048.

\noindent
\textit{Inference.}
We use FLUX's default rectified flow sampler~\cite{RFSampler} with 50 sampling steps.
FLUX computes timestep shifting based on the number of tokens.
Images at 1024$\times$2048 resolution surpasses its maximum number of tokens (4096), we thus uses its cutoff shifting value of $\mathrm{exp}(1.15) \approx 3.16$.
We tried larger value but observed degradation in the result.
We apply CFG on both text and image similar to~\cite{InstructPix2Pix}, with a scale of $2.0$ on text and $1.5$ on image.

\heading{Video model.}
We fine-tune Wan2.1-14B~\cite{Wan2.1}, a popular open source text-to-video diffusion transformer model.
Most of the model design is the same as the image model.
The only difference is that, in 3D RoPE, we offset the time dimension index of perspective tokens by 0.1 rather than 1 to avoid confusion with tokens from subsequent frames.
We fine-tune the entire model using the Adam optimizer with a learning rate of $1\times\text{10}^{-\text{5}}$, and a batch size of 64.
The same warmup schedule, gradient clipping, and CFG dropping are applied.
The model is first trained on ERP videos with 81 frames, 256$\times$512 resolution from the coarse subset for 10k steps, and then on ERP videos with 81 frames, 512$\times$1024 resolution from the high-quality subset for another 10k steps.

\noindent
\textit{Inference.}
We run 50 sampling steps with Wan's default sampler and timestep shifting of $3.0$, which outperforms $2.0$ and $5.0$ in our ablation.
We use a CFG weights of 3.0 for text and 2.0 for conditioning on perspective video.

\subsection{Evaluation Setup}
\label{app-subsec:eval-setup}

\heading{Perspective-to-360$^\circ\!$ image generation.}
We evaluate on the Laval Indoor~\cite{LavalIndoorDataset} and SUN360~\cite{SUN360Dataset} datasets.
To measure visual quality, we report Fréchet Inception Distance (FID)~\cite{FID}, Kernel Inception Distance (KID)~\cite{KID}, FID on CLIP~\cite{CLIP} features (CLIP-FID), and FID on features of an auto-encoder fine-tuned on panorama images (FAED)~\cite{PanFusion}.
In line with CubeDiff, FID, KID, and CLIP-FID are computed on 10 perspective crops (10 azimuth angles randomly sampled from [90$^\circ$, 270$^\circ$] to avoid overlap with the input view whose elevation=0$^\circ$) from the generated panorama using the \href{https://github.com/GaParmar/clean-fid}{clean-fid} package~\cite{CleanFID}.
Meanwhile, FAED is computed directly on the entire panorama as it measures the overall geometry of the ERP.
We adopt the implementation from PanFusion~\cite{PanFusion}\footnote{\url{https://github.com/chengzhag/PanFusion}}.
We also report CLIP-score (CS)~\cite{CLIPScore} between captions and the ERP images for text alignment.

\heading{Perspective-to-360$^\circ\!$ video generation.}
We follow Argus and use a hold out set of 101 videos as evaluation data.
However, as the exact eval split from Argus is unavailable, we reproduce the eval set based on offline communication with the authors.
We tested Argus on this reproduced eval set and ensured that the metrics are comparable to those reported in the paper.
Following Argus, we use two types of camera trajectories, namely, \textit{simulated} and extracted from \textit{real-world} videos, to obtain conditioning videos.
To measure fidelity, we report PSNR and LPIPS~\cite{LPIPS} between ground-truth and generated panorama videos within regions covered by the perspective video.
Concretely, it is computed by projecting a mask to the ERP space using ground-truth camera information at each step, and then take a union over all steps.
We also report FVD~\cite{FVD} on the ERP videos to measure the overall geometry and visual quality.
Finally, we adopt Imaging Quality, Aesthetic Quality, and Motion Smoothness from VBench~\cite{VBench} to evaluate overall generation quality.
VBench metrics are computed on perspective projections (front, left, right, back) of the generated panorama videos.

\begin{table}[!t]
\centering
\footnotesize
\renewcommand{\arraystretch}{1.1}

\begin{minipage}[t]{0.48\linewidth}
\centering
\caption{
\textbf{Base model and the parameter count (\#Params) of perspective-to-360$^\circ\!$ image generation methods.}
}
\label{tab:img-model-size}
\vspace{\tablecapmargin}
\begin{tabular}{lcc}
    \toprule
    \textbf{Method} & Base Model & \#Params \\
    \midrule
    OmniDreamer & VQ-GAN~\cite{VQGAN} & 2B \\
    PanoDiffusion & LDM~\cite{LDM} & 1.5B \\
    Diffusion360 & LDM~\cite{LDM} & 1B \\
    CubeDiff & LDM~\cite{LDM} & 1B \\
    \textbf{\MethodName} & FLUX.1~\cite{FLUX} & 12B \\
    \bottomrule
\end{tabular}
\end{minipage}
\hfill
\begin{minipage}[t]{0.5\linewidth}
\centering
\caption{
\textbf{Base model and the parameter count (\#Params) of perspective-to-360$^\circ\!$ video generation methods.}
}
\label{tab:video-model-size}
\vspace{\tablecapmargin}
\begin{tabular}{lcc}
    \toprule
    \textbf{Method} & Base Model & \#Params \\
    \midrule
    Imagine360 & AnimateDiff~\cite{AnimateDiff} & 3B \\
    Argus & SVD~\cite{SVD} & 1.5B \\
    ViewPoint & Wan2.1~\cite{Wan2.1} & 1.3B \\
    \textbf{\MethodName} & Wan2.1~\cite{Wan2.1} & 14B \\
    \bottomrule
\end{tabular}
\end{minipage}

\vspace{\tablemargin}
\end{table}

\subsection{Baselines}
\label{app-subsec:baseline-implementation}

\heading{Perspective-to-360$^\circ\!$ image generation.}
Since we follow the evaluation setup of the previous state-of-the-art (CubeDiff), we borrow the number of baselines from their paper.
Here, we only discuss CubeDiff.
CubeDiff leverages a cubemap representation, which projects the panorama image to six perspective views (six faces of a cube), each with 90$^\circ\!$ FoV.
At inference time, it inputs the conditioning image as the front face, and denoises the other five faces jointly.
This means they always assumes the input image has a 90$^\circ\!$ FoV and lies at the \textit{center} of the generated panorama, preventing them from adapting to in-the-wild images with arbitrary FoV and camera orientation.

\heading{Perspective-to-360$^\circ\!$ video generation.}
We compare with three recent works with released code.
Notably, if the method requires camera metadata to project the perspective video to the ERP space, we always use ground-truth camera metadata.
This provides additional information compared to \MethodName.

\begin{itemize}
\item \noindent\textit{Imagine360}~\cite{Imagine360} leverages a dual-branch architecture based on AnimateDiff~\cite{AnimateDiff} to process perspective and ERP data, connected with spherical attention.
We leverage their \href{https://github.com/3DTopia/Imagine360}{official code}.

\item \noindent\textit{Argus}~\cite{Argus} projects the perspective video to the ERP space, and fine-tunes SVD~\cite{SVD} to outpaint the entire panorama video.
We leverage their \href{https://github.com/Red-Fairy/argus-code}{official code}.

\item \noindent\textit{ViewPoint}~\cite{ViewPointPanoVideo} is inspired by the cubemap representation, and designs a viewpoint map representation with less spherical distortion compared to ERP.
It then fine-tunes Wan2.1~\cite{Wan2.1} on this new representation.
However, similar to CubeDiff, ViewPoint also places the conditioning video at the front view, leading to severely rotated panoramas when the input video has large camera motion.
We leverage their \href{https://github.com/ali-vilab/ViewPoint}{official code}.
\end{itemize}

\begin{table}[t]
\vspace{-0.5mm}
\caption{
\textbf{Perspective-to-360$^\circ\!$ image generation results at extreme pitch and roll angles on Laval Indoor.}
FoV is fixed at 90$^\circ\!$.
}
\vspace{\tablecapmargin}
\vspace{-0.5mm}
\label{tab:extreme-camera-pose}
\centering
\scriptsize
\setlength{\tabcolsep}{6pt}
\renewcommand{\arraystretch}{1.1}
\begin{tabular}{cc|cccc}
    \toprule
    \multirow{2}{*}{\textbf{Metric}} & \multicolumn{5}{c}{\textbf{Conditioning View} (Pitch, Roll)} \\
    \cmidrule(lr){2-6}
    & (0, 0) & (60, 0) & (-60, 0) & (60, 15) & (-60, -15) \\
    \midrule
    \textbf{FID} $\downarrow$ & 8.0 & 9.8 & 10.7 & 10.2 & 11.1 \\
    \midrule
    \textbf{FAED} $\downarrow$ & 9.8 & 10.5 & 10.8 & 10.7 & 11.0 \\
    \bottomrule
\end{tabular}
\vspace{\tablemargin}
\vspace{1mm}
\end{table}

\begin{figure*}[!t]
  \centering
    \includegraphics[width=\textwidth]{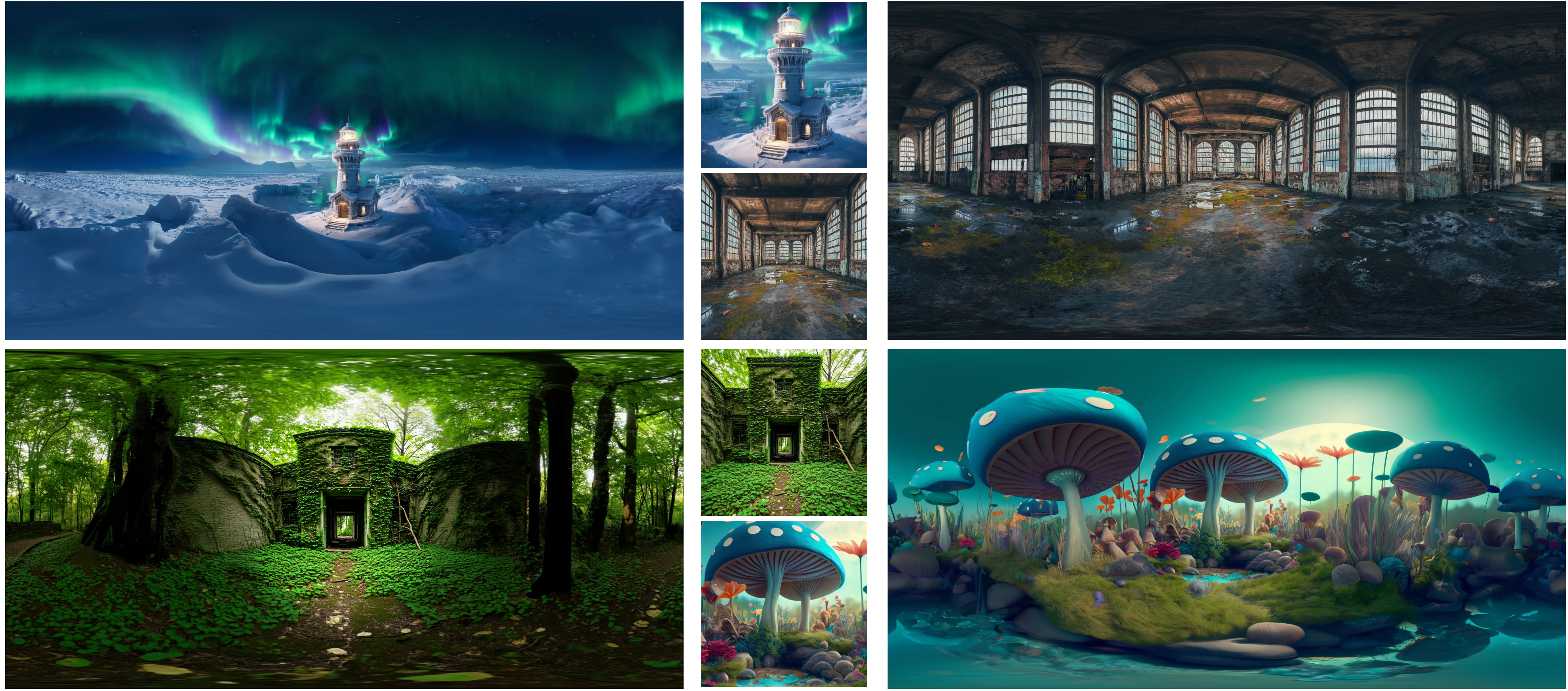}
    \vspace{\figcapmargin}
    \caption{
    \textbf{Panorama image generation results on out-of-distribution images.}
    The conditioning perspective images are shown at the middle.
    }
  \label{app-fig:qualitative-results-image}
  \vspace{\figmargin}
\end{figure*}

\section{More Experimental Results}
\label{app-sec:more-exp-results}

\subsection{Failure Cases of Channel-Concatenation Baselines}
\label{app-subset:channel-concat-failure}

\begin{figure*}[!t]
  \vspace{-2mm}
  \centering
    \includegraphics[width=\textwidth]{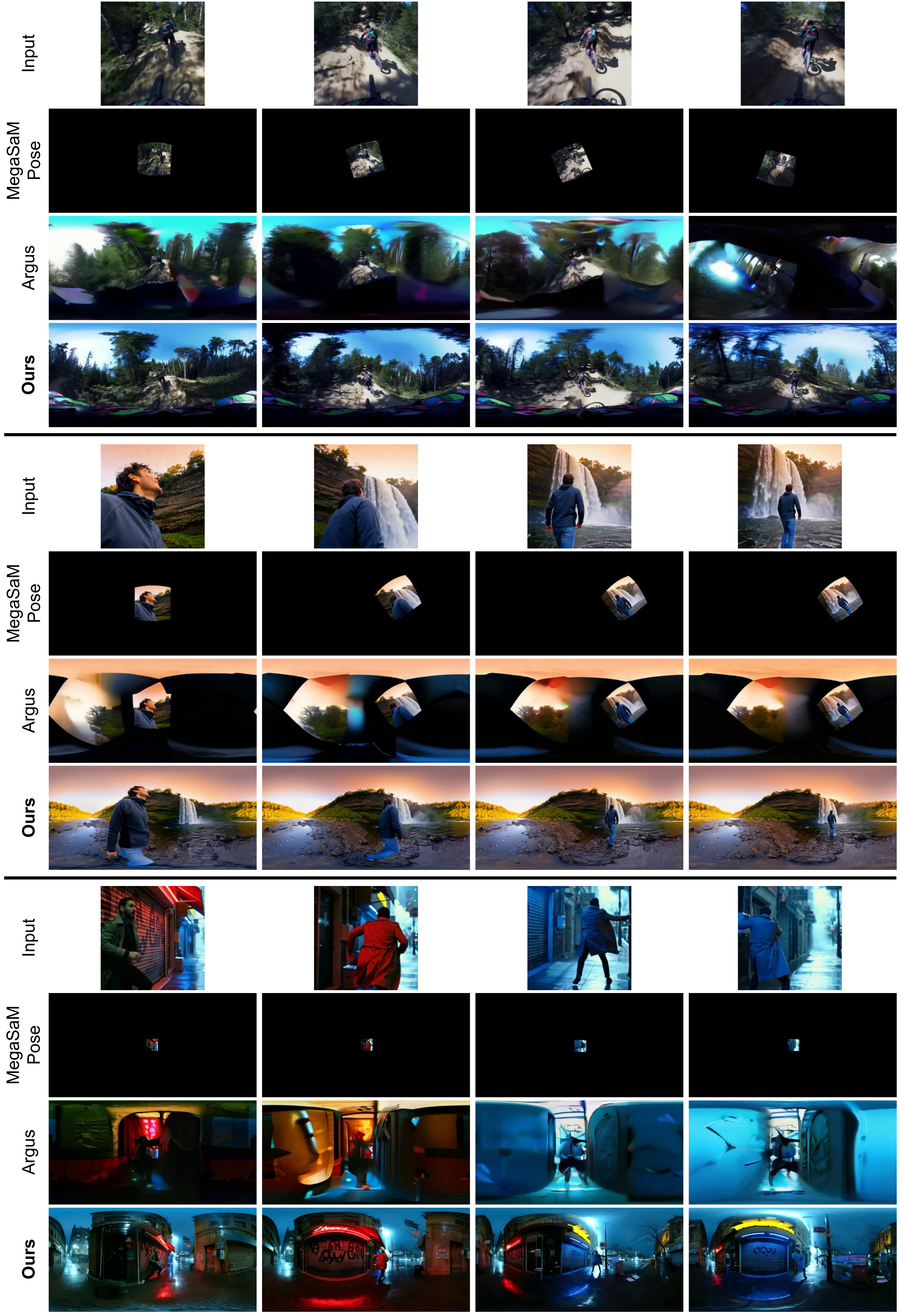}
    \caption{
    \textbf{Perspective-to-360$^\circ\!$ video generation on challenging input videos.}
    MegaSaM~\cite{MegaSaM} fails to predict the correct camera poses (first two examples) or FoV (last example), leading to degraded generation results from Argus.
    In contrast, our method runs end-to-end without a projection stage and thus generalizes well.
    }
  \label{app-fig:megasam-failure}
\end{figure*}

A popular line of work in perspective-to-360$^\circ\!$ generation first projects the perspective input to the ERP space, and then concatenates it with the noisy target latent channel-wise as model input.
This requires external models to estimate the FoV and camera orientations for the projection, which is tedious from a user perspective.
Moreover, channel-concatenation models may suffer from mistakes made by off-the-shelf camera estimators.
As shown in \Cref{app-fig:megasam-failure}, when the input video has complex camera trajectories or lighting conditions, even SoTA camera estimators like MegaSaM ~\cite{MegaSaM} fail.
In the first two examples, the predicted roll angles drift significantly, leading to severely tilted objects in the conditioning view.
In the last example, the predicted FoV is too small, making the projected view uninformative.
Argus is unable to correct the out-of-distribution conditioning input, and generates broken results.
In contrast, \MethodName gets rid of explicit camera information with a sequence-concatenation mechanism, and can still generate reasonable panorama videos from these challenging videos.

\begin{figure*}[!t]
  \centering
    \includegraphics[width=\textwidth]{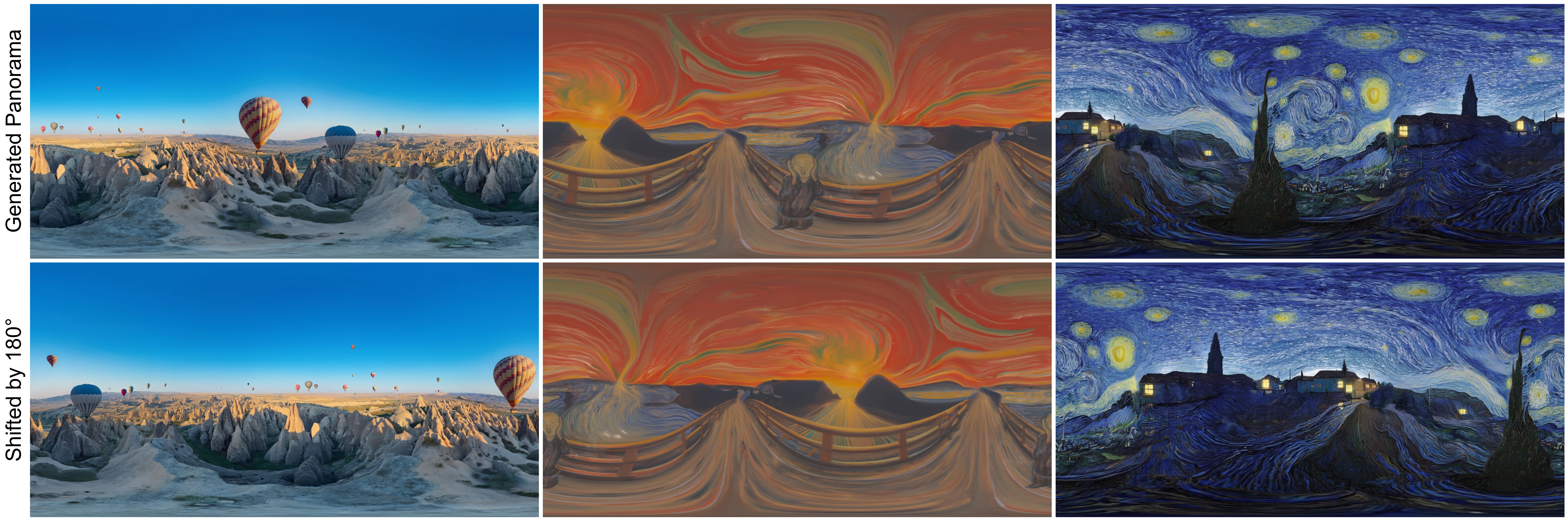}
    \vspace{\figcapmargin}
    \caption{
    \textbf{Visualization of boundary consistency.}
    We shift the generated panorama images by 180$^\circ\!$, repositioning the left and right image boundaries to the center.
    The smooth transition at the center demonstrates our seam-free generation.
    Please visit our \href{https://360anything.github.io/\#360-image-result}{project page} to explore more panorama images interactively.
    }
  \label{app-fig:image-seam}
  \vspace{\figmargin}
\end{figure*}

\subsection{Perspective-to-360$^\circ\!$ Image Generation}
\label{app-subset:more-pano-image-results}

\heading{Results on more extreme camera poses.}
\Cref{tab:extreme-camera-pose} evaluates \MethodName with pitches at $\pm$60$^\circ\!$ and rolls at $\pm$15$^\circ\!$.
The performance remains competitive in both metrics, showing our model's robustness at extreme camera poses.

\heading{Results on OOD images.}
We present results on out-of-distribution perspective images in \Cref{app-fig:qualitative-results-image}.
The conditioning images are generated by text-to-image models.
Despite being mainly trained on indoor synthetic data, \MethodName still generalizes to these OOD samples with high visual quality and correct overall structure.
\Cref{app-fig:image-seam} verifies that our generated panoramas are free of seams.

\begin{figure*}[!t]
  \vspace{3mm}
  \centering
    \includegraphics[width=\textwidth]{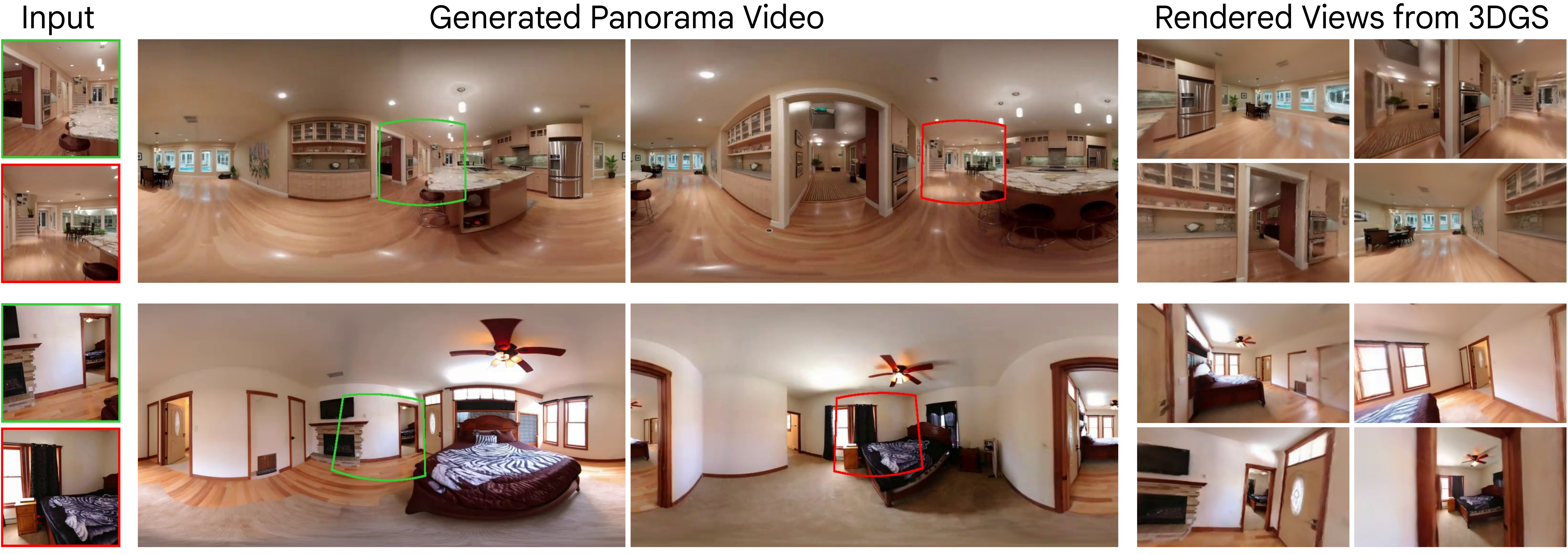}
    \vspace{\figcapmargin}
    \caption{
    \textbf{Qualitative results of 3D scene reconstruction.}
    Given an input monocular video (left), \MethodName outpaints the whole 360$^\circ\!$ viewpoint (middle), from which we can optimize a 3DGS (right).
    This allows fly-through exploration of the entire 3D scene.
    }
  \label{app-fig:3dgs-qualitative}
  \vspace{\figmargin}
\end{figure*}

\subsection{Perspective-to-360$^\circ\!$ Video Generation}
\label{app-subset:more-pano-video-results}

\begin{figure*}
\vspace{-2mm}
  \centering
  \begin{subfigure}[b]{1.0\textwidth}
    \includegraphics[width=\textwidth]{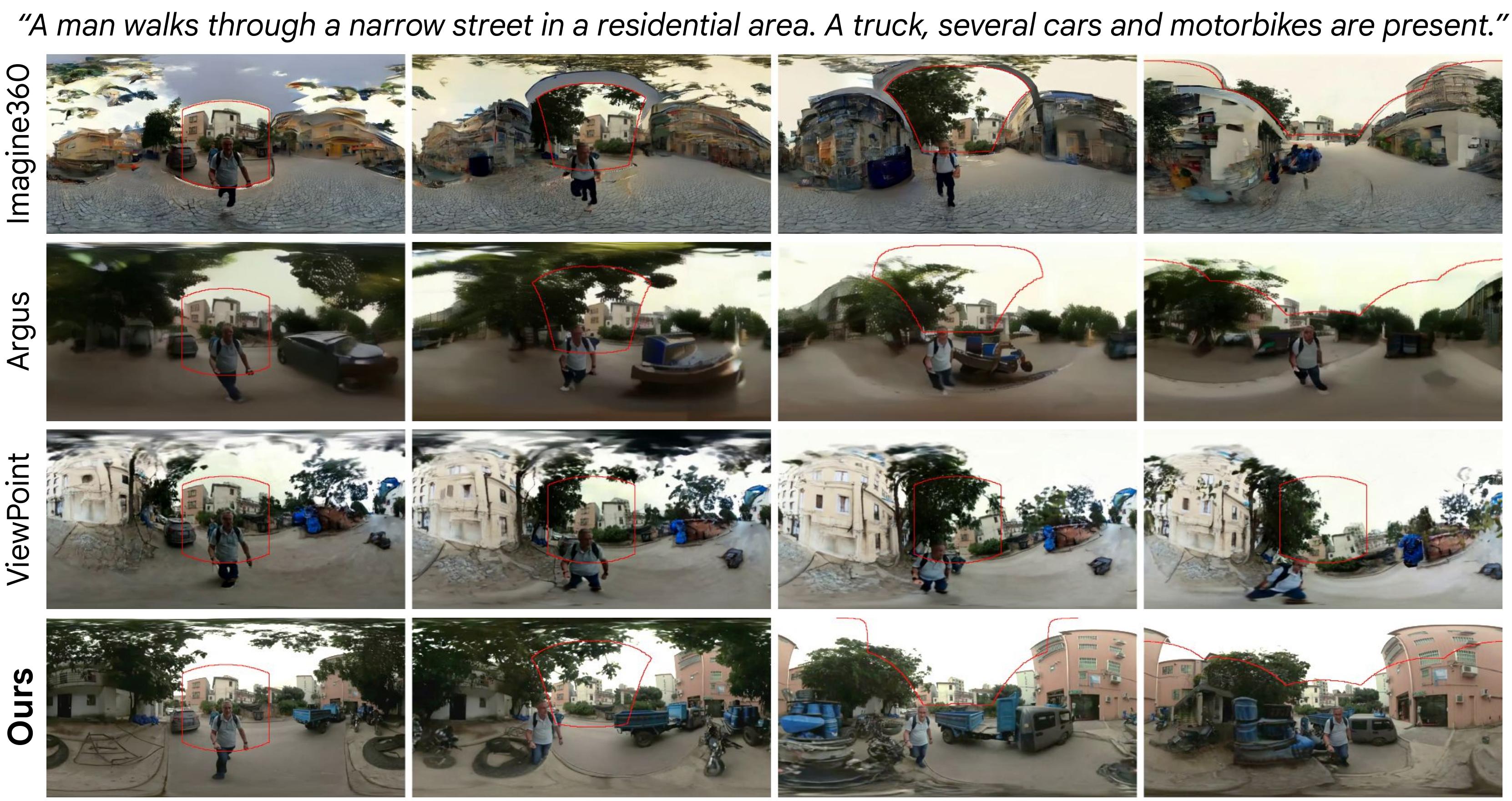}
  \end{subfigure}
  \begin{subfigure}[b]{1.0\textwidth}
    \includegraphics[width=\textwidth]{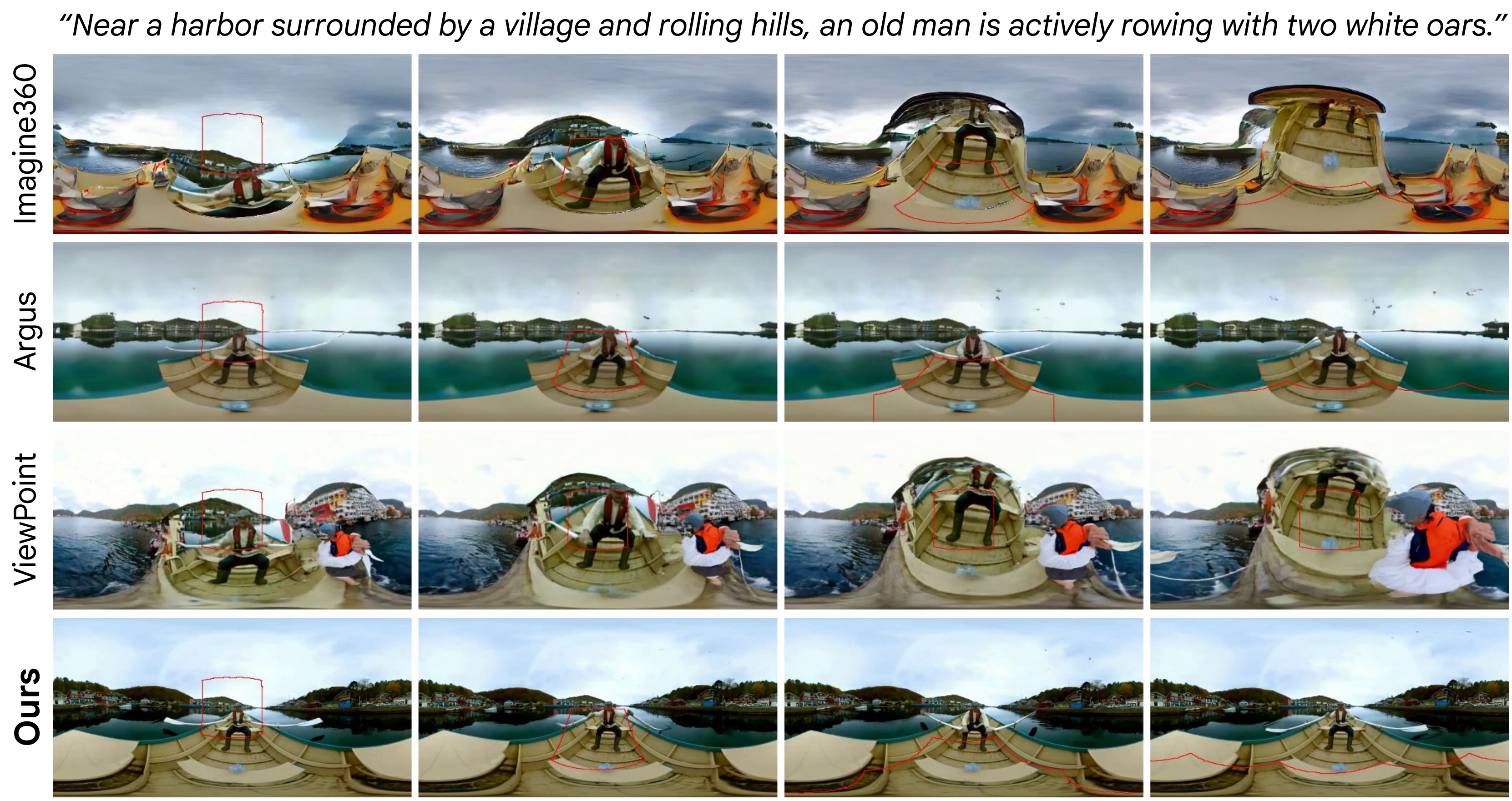}
  \end{subfigure}
  \begin{subfigure}[b]{1.0\textwidth}
    \includegraphics[width=\textwidth]{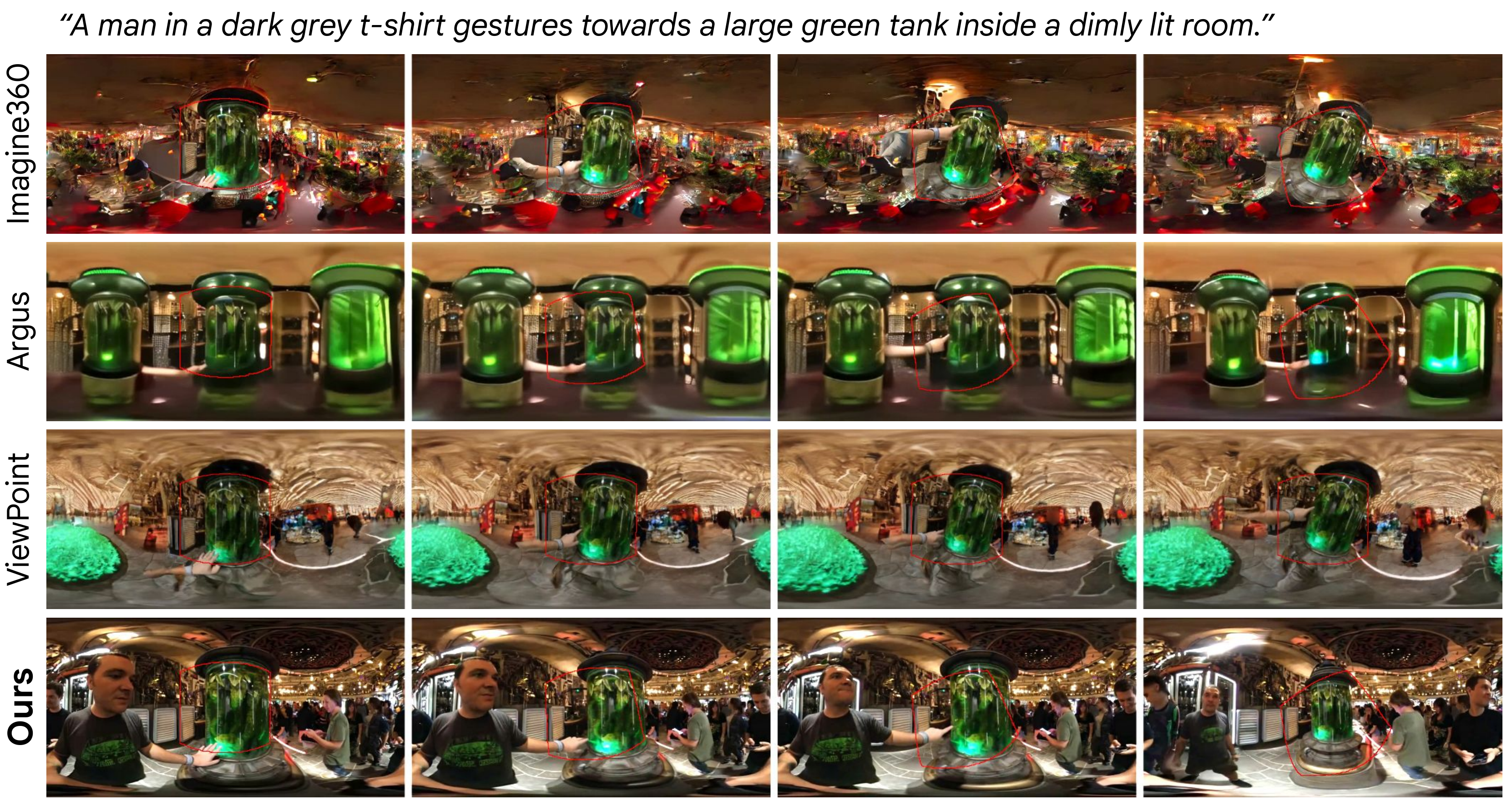}
  \end{subfigure}
  \caption{
    \textbf{Qualitative comparisons of perspective-to-360$^\circ\!$ video generation.}
  }
  \label{app-fig:qualitative-results-video}
\end{figure*}

\heading{Panoramic video outpainting.}
We show more qualitative comparisons in \Cref{app-fig:qualitative-results-video}.
In the first two examples, the conditioning perspective videos have large elevation changes (moves up and down).
Imagine360 and Argus handle this by projecting the perspective view to the ERP space as model conditioning.
Yet, they still generate videos with low visual quality due to the use of poor video backbones~\cite{AnimateDiff, SVD}.
ViewPoint uses the same Wan~\cite{Wan2.1} backbone as us.
However, it always treats the input frames as the front-view of the cubemap representation, and thus has to generate significantly rotated panoramas.
This leads to severely distorted humans and objects.
Thanks to the canonicalization training objective, \MethodName produces videos that are always upright, maintaining the correct scene geometry.
In the last example, the conditioning video contains a partially observed hand.
All baselines fail to outpaint the entire person, and only generate a small part of an arm.
In contrast, \MethodName outpaints the entire person following the text prompt, and still maintains it when the hand moves out of the input view.
This shows the strong world knowledge of our method.

\begin{figure*}[t]
  \centering
    \includegraphics[width=\textwidth]{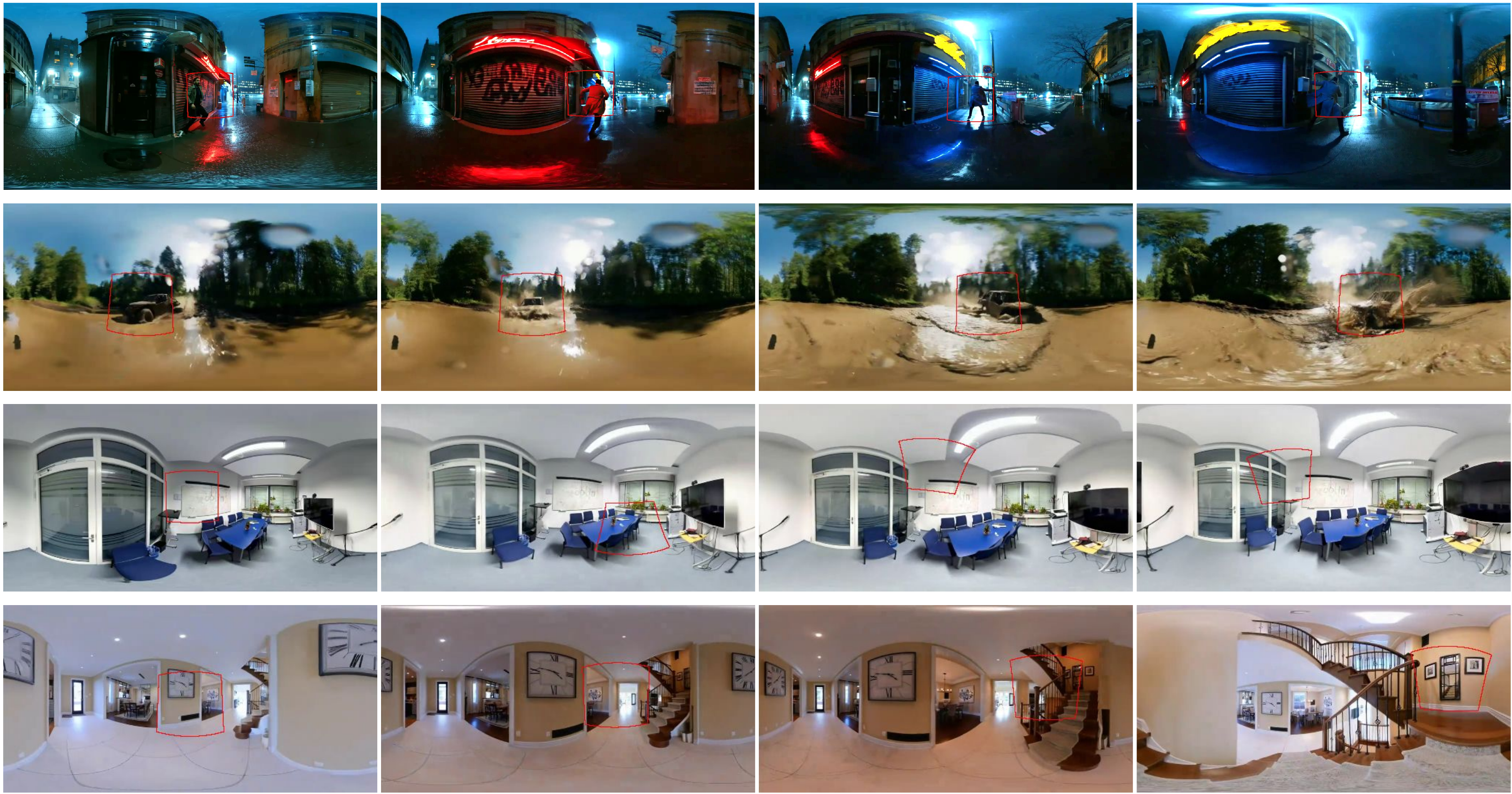}
    \vspace{\figcapmargin}
    \caption{
    \textbf{Panorama video generation given large motion videos.}
    Regions corresponding to the input conditioning video are highlighted in {\color{red} \textbf{red}}.
    We test on perspective videos with large object or camera motion.
    }
  \label{app-fig:qualitative-results-video-large-motion}
  \vspace{\figmargin}
\end{figure*}

\begin{figure*}[!t]
  \centering
    \includegraphics[width=\textwidth]{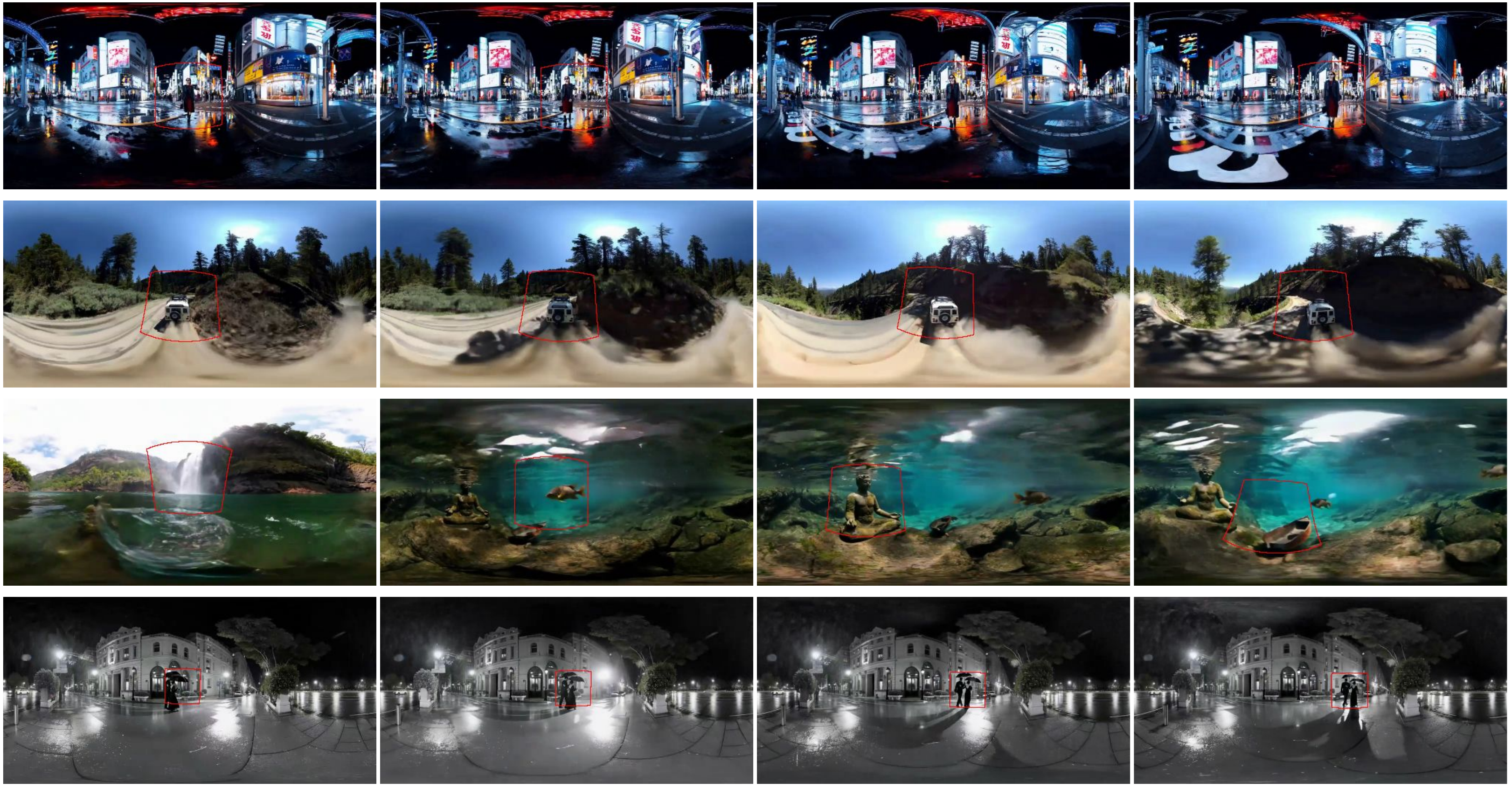}
    \vspace*{\figcapmargin}
    \caption{
    \textbf{Panorama video generation on AI generated videos.}
    Regions corresponding to the input conditioning video are highlighted in {\color{red} \textbf{red}}.
    We test on perspective videos generated by other video models.
    }
  \label{app-fig:qualitative-results-video-ood}
  \vspace{\figmargin}
\end{figure*}

\heading{3D scene reconstruction.}
We show more qualitative results in \Cref{app-fig:3dgs-qualitative}.
Given a narrow field-of-view video from RealEstate10K (RE10K)~\cite{RealEstate10k}, \MethodName synthesizes the entire 360$^\circ\!$ view of the room.
We can then train a 3D Gaussian Splatting model~\cite{3DGS} on the generated panoramas for novel view synthesis.

For quantitative evaluation, we seek metrics that do not require ground-truth because RE10K does not provide novel view images.
Instead, we take the mean reprojection error and image registration rate from COLMAP~\cite{Colmap} output to measure the quality of 3D reconstruction.
We run COLMAP on 50 input perspective videos and panorama videos generated from them (projected to four side views facing left, front, right, and back).
The input videos yield a baseline reprojection error of \textbf{0.61} pixels, representing the inherent noise in the SfM process on ground-truth data.
Our generated videos achieve an error of \textbf{0.74}.
While slightly higher than ground-truth, the sub-pixel error still suggests a good reconstruction.
For image registration rate, perspective videos achieve 94.5\%, while our model achieves a perfect 100\%, i.e., it does not drop any frames.
We attribute this improvement to our outpainted 360$^\circ\!$ field of view.
The expanded view coverage provides a broader set of geometric constraints and increased visual parallax, allowing the reconstruction pipeline to successfully incorporate frames that might otherwise fail to register in a narrower perspective view.
Overall, this demonstrates the 3D consistency of our generated panorama videos.

\heading{Results on large motion videos.}
We stress test our model on perspective videos with large object or camera motion.
\Cref{app-fig:qualitative-results-video-large-motion} shows that \MethodName is still able to produce temporally consistent videos that are in a stable canonical pose.
This shows that our model understands complex geometry and is able to find correspondence across the 4D world.

\heading{Results on OOD videos.}
We test our model on perspective videos generated by other video models, including Wan, Sora, Veo, and Runway Gen-4.5.
\Cref{app-fig:qualitative-results-video-ood} shows that \MethodName still generates panorama videos with high quality.
Notice how the reflection of billboards on water is well handled in the first example, even though it is not present in the input.
In the second example, the dust caused by the car persists after the car drives by.
In the third example, the model generates the statue underwater when the camera is above the water surface.
The last example features stylized black-and-white footage.

\begin{table}[t]
\caption{
\textbf{Comparison of \MethodName and its channel concatenation variant.}
We report VBench metrics and the inference latency when generating 81-frame, 512$\times$1024 panorama videos on one NVIDIA A100 GPU.
}
\vspace{\tablecapmargin}
\label{tab:runtime}
\centering
\scriptsize
\setlength{\tabcolsep}{6.0pt}
\renewcommand{\arraystretch}{1.1}
\begin{tabular}{cccc|cc}
    \toprule
    \textbf{Method} & Imag.$\uparrow$ & Aes.$\uparrow$ & Motion$\uparrow$ & VRAM & Runtime \\
    \midrule
    Channel-Concat & 0.5403 & 0.5265 & 0.9883 & 66 GB & 46 min \\
    \textbf{Sequence-Concat~(ours)} & \textbf{0.5515} & \textbf{0.5427} & \textbf{0.9885} & 76 GB & 55 min \\
    \bottomrule
\end{tabular}
\vspace{\tablemargin}
\end{table}

\heading{Ablation on conditioning mechanisms.}
We trained a variant of \MethodName that uses \emph{channel concatenation} to condition on the perspective video.
\Cref{tab:runtime} shows that we outperform it despite having the same model size, proving the effectiveness of our sequence concatenation technique.

Since the panorama data has 8$\times$ more tokens than the perspective data, the sequence concatenation method only increases the input sequence length by 1.125$\times$ compared to channel concatenation.
\Cref{tab:runtime} shows that the sequence concatenation version of \MethodName only introduces less than 20\% overhead.

\begin{figure*}[!t]
  \centering
    \includegraphics[width=\textwidth]{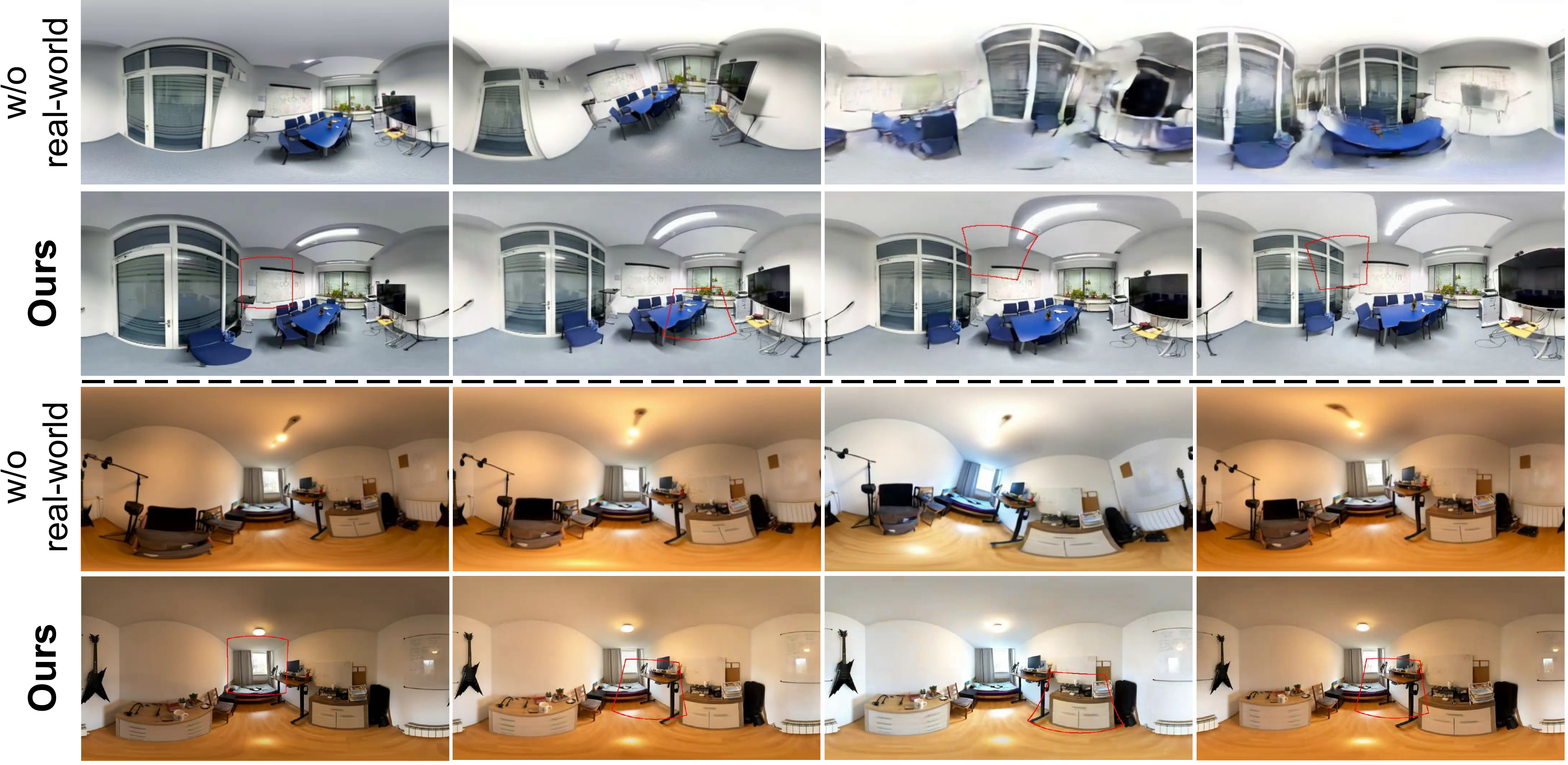}
    \vspace{\figcapmargin}
    \caption{
    \textbf{Ablation on real-world camera trajectories.}
    \MethodName uses both simulated and real-world camera trajectories to crop perspective videos in model training.
    Models without this setup generate panorama frames with changing gravity directions, fail to produce canonicalized videos and may suffer from broken structure.
    }
  \label{app-fig:ablate-real-world-cam}
  \vspace{\figmargin}
\end{figure*}

\heading{Ablation on real-world camera trajectories.}
We tested the model on in-the-wild perspective videos with large camera motion.
Large camera motion makes maintaining a canonicalized panorama output more challenging.
As shown in \Cref{app-fig:ablate-real-world-cam}, models trained only with simulated camera trajectories fail to produce frames with changing gravity directions.
In contrast, \MethodName with real-world camera trajectories generates stably canonicalized panorama videos.

\begin{figure*}[!t]
  \centering
    \includegraphics[width=\textwidth]{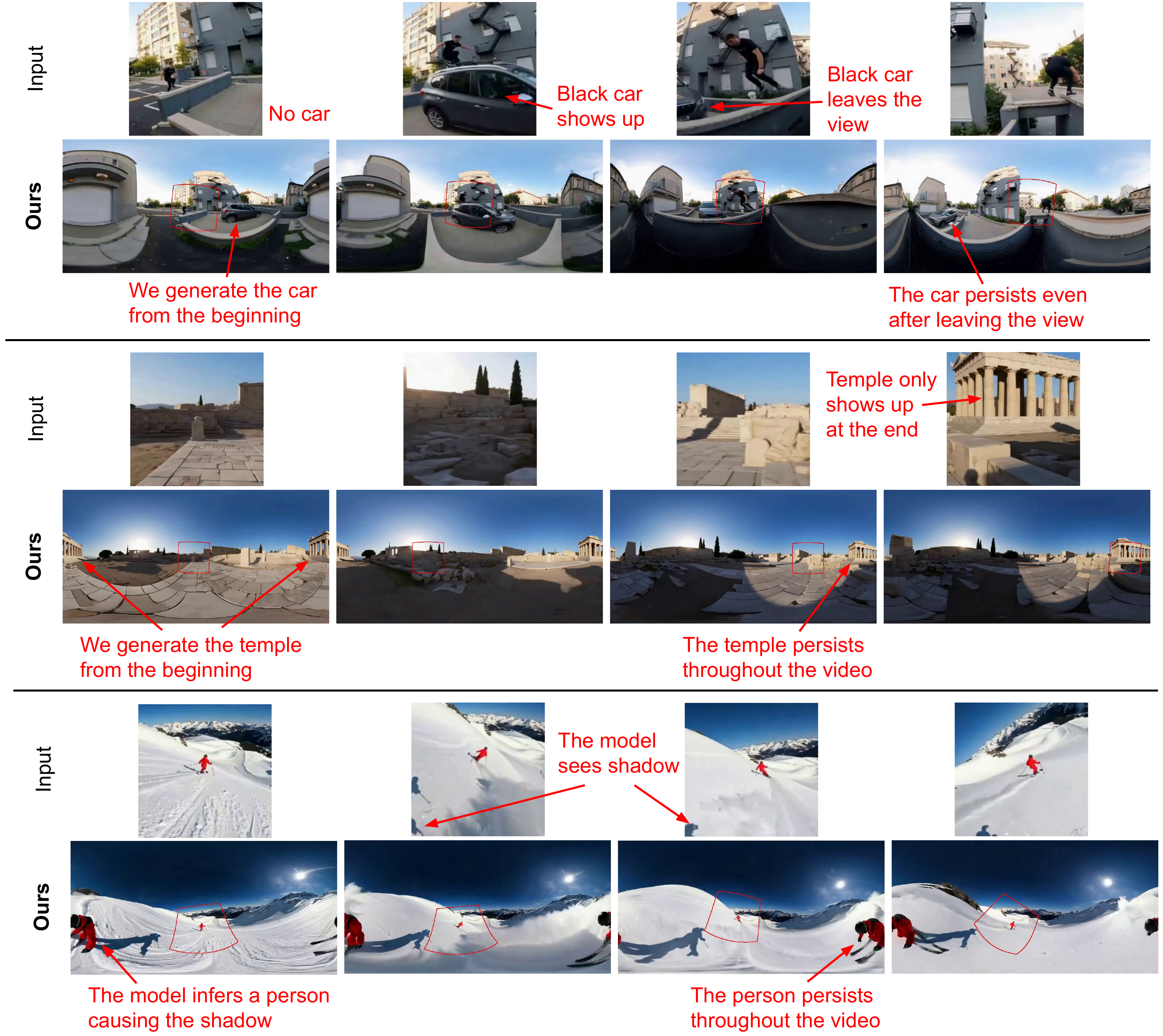}
    \caption{
    \textbf{Results on object permanence.}
    \MethodName consistently generates objects outside the perspective view in the 360$^\circ\!$ panorama (first two examples).
    It can also infer objects behind the camera based on cues such as shadows in the input view, even if the objects themselves never enter the perspective view (the person in the last example).
    }
  \label{app-fig:obj-permanence}
  \vspace{\figmargin}
\end{figure*}

\section{Limitations and Future Work}
\label{app-sec:limitations}

\MethodName is fine-tuned from a \textit{pre-trained} video diffusion model, and thus we are bounded by the capacity of the base model.
For example, it is challenging to outpaint scenes involving complex physics.
In addition, we also inherit the bias in its training data.
For example, the model sometimes generates panorama videos with black borders or undesired objects (e.g., a tripod or a human's hand) at the bottom of the video since they are common in YouTube 360$^\circ\!$ videos.

Due to the high resolution of panorama data (an ERP video has 8$\times$ number of pixels compared to a normal perspective video) and the limited compute, our current video model can only handle videos with 81 frames at 512$\times$1024 resolution.
A larger context window will enable larger-scale and more detailed 3D world generation.
To obtain higher-resolution panoramas, we tried existing video upsamplers designed for perspective videos~\cite{GoogleVeo}.
However, this often re-introduces seams at the ERP boundary, and distorts the structure of the ERP space, calling for research in panorama upsampling techniques.
To obtain a longer temporal horizon, we might combine \MethodName with recent progress in long video generation that distills bi-directional DiTs to \textit{causal} autoregressive DiTs~\cite{DiffusionForcing, CausVid, SelfForcing}.
Besides, the recent HY-World 2.0~\cite{HYWorld2.0} leverages panoramas for world generation.
The spatiotemporal memory mechanism can be adapted for long video generation as well.
In addition, the context mechanism in CubeComposer~\cite{CubeComposer} is inspiring for conditioning on perspective data in an autoregressive setup.

Standard video models struggle with object permanence--if a camera pans away and comes back, objects may change appearance or even disappear~\cite{GEN3C, VMem}.
In contrast, generating 360$^\circ\!$ panoramas forces the model to synthesize the entire scene simultaneously.
This serves as a ``working memory'' for the model, ensuring that out-of-sight regions remain consistent (\cref{app-fig:obj-permanence}).
An interesting direction is extending this concept to long-term ``episodic memory''~\cite{VideoMemory}.
Given the high spatio-temporal redundancy of panoramic videos, complex scenes can be efficiently compressed into a sequence of sparse panoramic keyframes.
However, modeling object occlusion within such sparse representations remains an open challenge.

\end{document}